\documentclass[10pt,twocolumn,letterpaper]{article}
\usepackage[accsupp]{axessibility}  
\usepackage[pagenumbers]{cvpr} 
\definecolor{cvprblue}{rgb}{0.21,0.49,0.74}
\usepackage[pagebackref,breaklinks,colorlinks,allcolors=cvprblue]{hyperref}

\usepackage{fancyhdr} 

\fancypagestyle{firstpage}{
    \fancyhf{}
    
    \fancyfoot[C]{\small 
        Accepted for publication in the IEEE/CVF Conference on Computer Vision and Pattern Recognition (CVPR), 2026. \\
        \textcopyright\ 2026 IEEE. Personal use permitted; other uses require IEEE permission.
    }
}

\usepackage{bbm}
\usepackage{multirow}
\newcommand{\ours}{HINGE}
\newcommand{\meansd}[2]{\ensuremath{#1_{\scriptstyle \pm #2}}}
\newcommand{\meansdbf}[2]{\ensuremath{\mathbf{#1_{\scriptstyle \pm #2}}}}
\newcommand{\meansdul}[2]{\ensuremath{\underline{#1_{\scriptstyle \pm #2}}}}
\def\paperID{14125} 
\def\confName{CVPR}
\def\confYear{2026}

\title{Adapting a Pre-trained Single-Cell Foundation Model to Spatial Gene Expression Generation from Histology Images} 

\author{
Donghai Fang$^{1,2}$ \quad Yongheng Li$^{2}$ \quad Zhen Wang$^{1}$\textsuperscript{*} \quad Yuansong Zeng$^{3}$\textsuperscript{*} \quad Wenwen Min$^{2}$\textsuperscript{*}\\
$^{1}$Sun Yat-sen University, China \quad
$^{2}$Yunnan University, China \quad
$^{3}$Chongqing University, China \\
{\tt\small wangzh665@mail.sysu.edu.cn, zengys@cqu.edu.cn, minwenwen@ynu.edu.cn}
}

\begin{document}

\maketitle
\thispagestyle{firstpage} 

\begingroup
\renewcommand\thefootnote{\fnsymbol{footnote}}
\footnotetext[1]{Corresponding authors}
\endgroup

\begin{abstract}
Spatial transcriptomics (ST) enables spot-level in situ expression profiling, but its high cost and limited throughput motivate predicting expression directly from H\&E-stained histology. 
Recent advances explore using score- or flow-based generative models to estimate the conditional distribution of gene expression from histology, offering a flexible alternative to deterministic regression approaches. 
However, most existing generative approaches omit explicit modeling of gene–gene dependencies, undermining biological coherence. 
Single-cell foundation models (sc\mbox{-}FMs), pre-trained across diverse cell populations, capture these critical gene relationships that histology alone cannot reveal. 
Yet, applying expression-only sc\mbox{-}FMs to histology-conditioned expression modeling is nontrivial due to the absence of a visual pathway, a mismatch between their pre-training and conditional ST objectives, and the scarcity of mixed-cell ST supervision. 
To address these challenges, we propose \textbf{HINGE} (\textbf{HI}stology-co\textbf{N}ditioned \textbf{GE}neration), which retrofits a pre-trained sc\mbox{-}FM into a conditional expression generator while mostly preserving its learned gene relationships. 
We achieve this by introducing \textit{SoftAdaLN}, a lightweight, identity-initialized modulation that injects layer-wise visual context into the backbone, coupled with an expression-space \textit{masked diffusion} objective and a warm-start curriculum to ensure objective alignment and training stability. 
Evaluated on three ST datasets, \ours\ outperforms state-of-the-art baselines on mean Pearson correlation and yields more accurate spatial marker expression patterns and higher pairwise co-expression consistency, establishing a practical route to adapt pre-trained sc\mbox{-}FMs for histology-conditioned spatial expression generation. 
\end{abstract}

\section{Introduction}
\label{sec:intro}
Spatial transcriptomics (ST) enables the measurement of gene expression in its native spatial context, but its high cost and limited throughput hinder widespread adoption~\cite{staahl2016visualization, neidlinger2025benchmarking}. 
A practical alternative is to infer spatial gene expression directly from Hematoxylin–Eosin (H\&E) histology (e.g., whole-slide images), which are routinely acquired~\cite{chen2024UNI, lu2024CONCH, chen2025omic}. 
The goal in this setting is to perform spot-level inference from histology, aiming for both accurate and spatially coherent predictions.

Most existing methods adopt a deterministic image-to-gene regression paradigm, mapping histology patches to predicted expression vectors at each spot~\cite{he2020stnet,pang2021hisTogene}. 
At the same time, biological variability, spatial heterogeneity, and measurement noise mean that the observed expression at a given spot is not uniquely determined by the local histology. 
Motivated by this, recent work explores score- or flow-based generative models that approximate the conditional distribution of gene expression given histology, providing a flexible alternative to standard regression approaches~\cite{zhu2025Stem, huang2025STFlow}. 
Methods such as Stem~\cite{zhu2025Stem} and STFlow~\cite{huang2025STFlow} instantiate this idea with histology-conditioned generative models that learn a distribution over spatial gene expression conditioned on histology images.

Despite these advances, current generative methods remain limited in a critical aspect: they omit explicit modeling of gene–gene dependencies—regulatory and co-expression patterns that are difficult to infer from histology alone but are essential for producing biologically coherent predictions. 
An emerging avenue to address this limitation is to leverage single-cell foundation models~\cite{baek2025scfms} (sc\mbox{-}FMs; e.g., scGPT~\cite{cui2024scscgpt}, scFoundation~\cite{hao2024large}, and CellFM~\cite{zeng2025cellfm}) pre-trained via masked autoencoding on large-scale single-cell RNA sequencing (scRNA-seq) across diverse cell populations, thereby encoding complex gene relationships that histology alone does not directly expose. 
Building on this premise, emerging work has begun exploring the transfer of sc\mbox{-}FM's knowledge to ST tasks~\cite{wang2025scgpt}.
However, this line of work primarily remains in expression space and offers limited sensitivity to histology, which further reflects the increasing difficulty of adapting expression\mbox{-}only sc\mbox{-}FMs to histology\mbox{-}conditioned expression modeling. 

Directly transferring sc\mbox{-}FMs to spatial gene expression generation presents four key challenges: 
\textit{(i) Modality gap.} sc\mbox{-}FMs are pre-trained exclusively in expression space and lack a visual pathway for histology, making cross-modal conditioning nontrivial~\cite{han2025reusability}. 
\textit{(ii) Objective mismatch.} sc\mbox{-}FMs are commonly trained with masked autoencoding~\cite{zeng2025cellfm, baek2025scfms, wang2025scgpt}, whereas most histology-to-expression methods adopt regression or DDPM-style denoising with all the input dimensions corrupted by Gaussian noise.
This input-and-supervision mismatch can hinder the transfer of pre-trained gene–gene patterns. 
\textit{(iii) Compositional shift.} Unlike scRNA-seq, which profiles individual cells, ST captures gene expression from local mixtures of cell types. 
This cross-omics discrepancy introduces expression shifts that complicate the reuse of single-cell models in ST contexts~\cite{kleshchevnikov2022cell2location, theodoris2023transfer}.
\textit{(iv) Limited supervision.} ST datasets are limited in size, and spot-level measurements are often noisy due to mixed-cell composition. These limitations make full\mbox{-}model fine\mbox{-}tuning prone to catastrophic forgetting of learned knowledge.  

To address these challenges, we propose \textbf{HINGE} (\textbf{HI}stology-co\textbf{N}ditioned \textbf{GE}neration), which retrofits a pre-trained sc\mbox{-}FM into a conditional expression generator while mostly preserving its learned gene relationships. 
Built on a pre-trained sc\mbox{-}FM instance, we keep its expression-only backbone frozen and install a lightweight conditioning pathway that provides a visual route from histology. 
To implement this pathway for effective conditional control while mitigating catastrophic forgetting for this sc\mbox{-}FM, we introduce \textbf{SoftAdaLN}, an identity-initialized layer-wise modulation that injects histology and timestep context throughout the backbone and keeps the sc\mbox{-}FM's original behavior at the beginning of fine-tuning.
In order to align with the masked autoencoding pre-training, we design an expression-space \textbf{masked diffusion} process in which the reverse transitions are parameterized by the histology-conditioned backbone and predictions progressively reveal masked gene entries. 
To better match the pre-training regime and stabilize early updates, we design a \textbf{warm-start curriculum} that samples low-mask timesteps during the initial fine-tuning steps.
Evaluated on three ST datasets from different tissues, \ours{} outperforms six state-of-the-art regression and generative baselines in mean Pearson correlation across genes and, consistent with prior generative studies~\cite{zhu2025Stem, huang2025STFlow}, yields more coherent spatial marker maps and higher pairwise co-expression consistency, providing a practical route to adapt sc\mbox{-}FMs for spatial expression generation. Code is available at: \textbf{\url{https://github.com/donghaifang/HINGE}}.

We summarize our main contributions as follows: 

\newlength{\labwd}\settowidth{\labwd}{\labelitemi\ }
\noindent\hangindent=\labwd\hangafter=1 \llap{\labelitemi\ }
We present \ours{}, the first framework to adapt pre-trained expression-only sc\mbox{-}FMs for histology-conditioned gene expression generation.

\noindent\hangindent=\labwd\hangafter=1 \llap{\labelitemi\ }
We introduce SoftAdaLN, masked diffusion, and a warm-start curriculum to enable effective and stable knowledge transfer, even under limited ST supervision.

\noindent\hangindent=\labwd\hangafter=1 \llap{\labelitemi\ }
\ours{} sets new state-of-the-art across three ST datasets, outperforming regression and generative baselines in accuracy, spatial coherence, and co-expression fidelity. 

\begin{figure*}[t]
  \centering
   \includegraphics[width=0.94\linewidth]{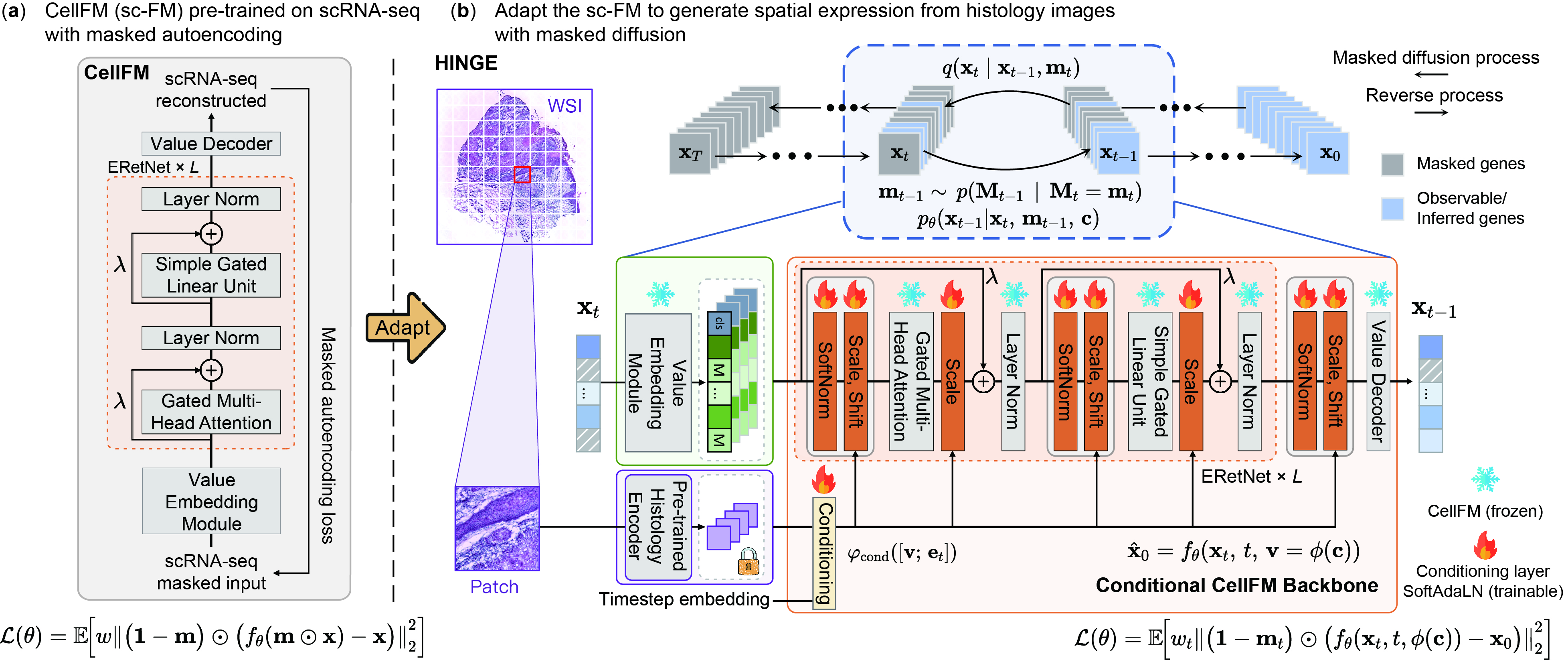}
   \caption{Overview of \ours. \textbf{(a)} Depicts the CellFM architecture, which is a single-cell foundation model (sc\mbox{-}FM) pre-trained on scRNA-seq with masked autoencoding. \textbf{(b)} In \ours{}, the conditional denoising model is instantiated from CellFM and augmented with identity-initialized SoftAdaLN that injects histology and timestep context into each transformer layer within a stochastic masked diffusion process. This design keeps the training objective aligned with CellFM’s masked autoencoding for coherence in ST, thereby largely preserving the gene relationships learned from scRNA-seq. }
   \label{fig2:model}
\end{figure*}

\section{Related Work}
\label{sec:Related}
\noindent\textbf{Histology-to-expression prediction.}
One line of work treats spatial gene expression prediction as deterministic regression from histology to expression. 
Early methods such as ST-Net~\cite{he2020stnet}, HisToGene~\cite{pang2021hisTogene} and Hist2ST~\cite{zeng202hist2st} use convolutional or transformer backbones to map histology to spatial expression. 
To capture broader context, subsequent models introduce multi-scale fusion: TRIPLEX~\cite{chung2024TRIPLEX} and MO2ST~\cite{wang2025m2ost} aggregate hierarchical image features spanning cellular patterns and larger tissue organization. 
Another line models spatial relationships between spots via graph reasoning. 
MERGE~\cite{ganguly2025merge} builds hierarchical graphs to propagate information across distant regions, HGGEP~\cite{li2024gene} employs hypergraphs to encode higher-order neighborhoods, and M2TGLGO~\cite{shi2025multigo} injects multimodal prior information into graph attention. 
Complementary contrastive approaches such as BLEEP~\cite{xie2023BLEEP} and mclSTExp~\cite{min2024multimodal} learn a joint embedding of histology and expression by enforcing consistency across nearby spots~\cite{lee2024pathomclip}. 

Overall, these frameworks span local encoders, multi-scale architectures, graph-structured models~\cite{yu2023stgcl,fatemi2023feasibility}, and contrastive designs, and have driven substantial progress in histology-based expression prediction. 
However, spot-level expression depends on underlying cell-type composition, cellular states, and microenvironmental factors that are only partially reflected in the visible histology, so the expression observed at a given spot cannot be fully determined from the surrounding tissue image alone. This leaves room for complementary formulations that go beyond a single point prediction from histology. 

\noindent\textbf{Histology-conditioned generative models.}
In addition to histology-to-expression regression models, recent work has explored generative models conditioned on histology images. 
Stem adopts a conditional diffusion model to sample spatial expression from histology, whereas STFlow employs flow matching to learn the joint slide-level expression distribution, with both models conditioned on tissue images~\cite{zhu2025Stem,ho2020DDPM,huang2025STFlow,lipman2022flow}. 
These approaches model a conditional distribution over spatial gene expression given histology images, but are typically trained without leveraging the gene--gene dependencies encoded in pre-trained sc\mbox{-}FMs, which are difficult to recover reliably from histology images alone, leaving open how to couple histology-conditioned generation with sc\mbox{-}FM pre-training.

\noindent\textbf{Single-cell knowledge transfer to spatial omics.}
Single-cell foundation models (sc\mbox{-}FMs) such as scFoundation~\cite{hao2024large}, scGPT~\cite{cui2024scscgpt}, and CellFM~\cite{zeng2025cellfm} are pre-trained with masked autoencoding–style objectives on large scRNA-seq corpora, and have been shown to capture gene--gene dependencies that are not straightforward to recover solely from histology images~\cite{baek2025scfms}. 
Recent work transfers this single-cell knowledge into spatial transcriptomics~\cite{yiu2025transformative}. 
Nicheformer~\cite{schaar2024nicheformer} and SToFM~\cite{zhao2025stofm} are spatial omics foundation models jointly pre-trained on spatial transcriptomics data, sometimes together with single-cell profiles, to embed single-cell information into spatial representations. 
In contrast, scGPT\mbox{-}spatial~\cite{wang2025scgpt} starts from a pre-trained scGPT and continually pre-trains it on large spatial corpora with spatially aware reconstruction, directly extending an expression-only sc\mbox{-}FM to spatial omics.  
These approaches demonstrate the promise of single-cell pre-training for spatial biology, primarily by enriching spatial representations in expression space. 
However, they do not address the setting of generating spatial gene expression directly from histology images by adapting a pre-trained expression-only sc\mbox{-}FM trained to histology-conditioned generation, leaving this type of cross-modal adaptation largely unexplored~\cite{luo2025deep}.

\section{Methodology}
\label{sec:Method}
In this section, we present \ours{}, which retrofits a pre-trained masked autoencoding sc\mbox{-}FM into a histology-conditioned generator for ST. 
We keep its expression-only backbone frozen and add a lightweight conditioning pathway that provides a visual route from histology. 
Throughout, we instantiate the backbone with CellFM (as shown in Fig.~\ref{fig2:model}(a)), while the architecture remains compatible with other masked-autoencoding sc\mbox{-}FMs (Sec.~\ref{sec:method:notation}). 
As shown in Fig.~\ref{fig2:model}(b), we introduce a masked diffusion process for histology-to-expression generation whose reverse steps are parameterized by a histology-conditioned CellFM and trained with an objective aligned with its masked autoencoding pre-training regime (Sec.~\ref{sec:method:masked_diffusion}). 
To enable conditional generation, we insert identity-initialized SoftAdaLN modulators into each transformer layer of CellFM, injecting histology and timestep signals so that the model leverages gene dependencies learned from scRNA-seq and mitigates compositional shift between the two transcriptomic settings (Sec.~\ref{sec:method:architecture}). 
Finally, a warm-start curriculum that initially samples low-mask timesteps stabilizes early training and further matches the pre-training regime (Sec.~\ref{sec:method:training}).

\subsection{Notation and Background}
\label{sec:method:notation}
Let $G$ denote the number of genes. Each spatial spot is associated with a gene-expression vector $\mathbf{x} \in \mathbb{R}^G$ and a corresponding histology image patch $\mathbf{c}$, where each component $x^{(g)}$ denotes the expression of the $g$-th gene at that spot.
Following the standard ST setting, we assume access to $N$ i.i.d.\ paired observations $\mathcal{D}=\{(\mathbf{x}_i,\mathbf{c}_i)\}_{i=1}^N$ drawn from an underlying joint distribution $p(\mathbf{X},\mathbf{C})$.

Our goal is to estimate the conditional distribution $p(\mathbf{X}\mid \mathbf{C})$, enabling prediction of spatial gene expression from histology while preserving intrinsic gene--gene dependencies.
Large single-cell foundation models such as CellFM are pre-trained to model the marginal distribution $p(\mathbf{X})$ from scRNA-seq data, without access to histology. These models typically use a masked autoencoding objective: a binary mask $\mathbf{m}\in\{0,1\}^G$ is drawn and applied to the input expression vector $\mathbf{x}$, and the model learns to reconstruct $\mathbf{x}$ from the masked one $\mathbf{m}\odot\mathbf{x}$, where $\odot$ denotes element-wise product.

For ST, generative approaches often discretize a stochastic process $\{\mathbf{X}_t\}_{t=0}^T$ in the continuous expression space, with $p(\mathbf{X}_0\mid \mathbf{C})=p(\mathbf{X}\mid \mathbf{C})$ and $p(\mathbf{X}_T\mid \mathbf{C})\approx \mathcal{N}(\mathbf{0},\mathbf{I}_G)$, where a forward process progressively corrupts $\mathbf{X}_0$ with Gaussian noise, and the reverse dynamics is parameterized by a denoising network. 
Sampling the expression for a given histology $\mathbf{c}$ means starting with $\mathbf{x}_{T} \sim \mathcal{N}(\mathbf{0},\mathbf{I}_G)$.
To transit from timestep $t$ to $(t-1)$, the denoising network receives $(\mathbf{x}_t, t, \mathbf{c})$ and produces an estimated clean sample $\hat{\mathbf{x}}_0=f_{\theta}(\mathbf{x}_t, t, \mathbf{v}=\phi(\mathbf{c}))$, where $f_{\theta}(\cdot)$ denotes the denoising network's backbone, and $\phi(\cdot)$ denotes a pre-trained histology encoder.

\subsection{Expression Modeling via Masked Diffusion}
\label{sec:method:masked_diffusion}
\ours\ aims to adapt a sc\mbox{-}FM for generative modeling of histology-to-expression mapping.
We chose CellFM for this purpose, given its strong capacity to capture gene dependencies.
As mentioned in Sec.~\ref{sec:method:notation}, CellFM is trained with a masked autoencoding objective, taking $\mathbf{m}\odot\mathbf{x}$ as its input, where a subset of components is masked to zero while the rest remain unchanged.
When retrofitting it into the generative approaches' backbone, it would take $\mathbf{x}_t$ as its input, which is often obtained by independently perturbing each component of $\mathbf{x}$ with Gaussian noise. 
This discrepancy in input distributions between masked autoencoding and diffusion-style denoising objectives can impede effective knowledge transfer. 
To bridge this gap, we introduce a dedicated stochastic masking process tailored for \ours. 

\noindent\textbf{Forward process.}
To obtain partially observed $\mathbf{x}_t$ (rather than perturbing every component), we augment $\{\mathbf{X}_t\}_{t=0}^{T}$ with $\{\mathbf{M}_{t}\}_{t=0}^{T}$ and define the transition probability distribution as \( q(\mathbf{X}_t,\mathbf{M}_t\mid \mathbf{X}_{t-1},\mathbf{M}_{t-1})= q(\mathbf{M}_t\mid \mathbf{M}_{t-1})\,\delta_{\mathbf{M}_t\odot \mathbf{X}_{t-1}}(\mathbf{X}_t) \), where $\delta_{x}(\cdot)$ denotes the Dirac delta function.
Given the current state $(\mathbf{m}_{t-1},\mathbf{x}_{t-1})$, this means first sampling $\mathbf{m}_{t}$ from $q(\mathbf{M}_{t}\mid \mathbf{M}_{t-1}=\mathbf{m}_{t-1})$, and then obtaining $\mathbf{x}_{t}$ by masking $\mathbf{x}_{t-1}$ with $\mathbf{m}_{t}$.

For each clean sample $\mathbf{x}_0$, we initialize the process with $\mathbf{m}_0 = \mathbf{1}$ deterministically.
Then, we define the transition of masks as $q(\mathbf{M}_t \mid \mathbf{M}_{t-1})
= \prod_{g=1}^G \Bigl[
\mathrm{Bern}\!\bigl(\mathbf{M}_{t}^{(g)} ; \mathbf{M}_{t-1}^{(g)}(1-p_t)\bigr)
\Bigr]$, where $\mathrm{Bern}(\cdot;p)$ represents a Bernoulli distribution that gives $1$ with probability $p$.
This formulation perturbs each component of the mask independently, and once a component becomes $0$, it remains $0$.
To ensure the mask ratio (i.e., the fraction of zeros) increases monotonically, we set the cumulative visibility using a power schedule \(\bar{\alpha}_t=\bigl(1-\tfrac{t}{T}\bigr)^{\zeta}\) with \(\zeta>0\), which induces per-step drop probabilities \(p_t=1-\bar{\alpha}_t/\bar{\alpha}_{t-1}\) for \(t=1,\ldots,T\).

As in vanilla diffusion models, our single-step transition definition allows efficient sampling of the state at timestep $t$ without simulating the entire process:
\begin{equation}
\begin{aligned}
    q(\mathbf{X}_t,\mathbf{M}_t\mid \mathbf{X}_{0},\mathbf{M}_{0}) &= q(\mathbf{M}_t\mid \mathbf{M}_{0})\,\delta_{\mathbf{M}_t\odot \mathbf{X}_{0}}(\mathbf{X}_t)\\
    q(\mathbf{M}_t \mid \mathbf{M}_{0}) &= \prod_{g=1}^G \Bigl[
\mathrm{Bern}\!\bigl(\mathbf{M}_{t}^{(g)} ; \bar{\alpha}_t \bigr)
\Bigr].
\end{aligned}
    \label{eq:fwd}
\end{equation}
It is easy to verify that as $t$ increases, $\bar{\alpha}_t$ approaches $0$, and thus the mask $\mathbf{m}_t$ gradually obscures $\mathbf{x}_t$ until all components are masked.

\noindent\textbf{Reverse process.}
The reverse process begins with $\mathbf{m}_{T} = \mathbf{0}$ and $\mathbf{x}_{T}=\mathbf{0}$, consistent with the masked diffusion models in the discrete domain.
At each time step $t=1,\ldots,T$, we factor the one-step reverse transition as \(p(\mathbf{X}_{t-1},\mathbf{M}_{t-1}\mid \mathbf{X}_t,\mathbf{M}_t,\mathbf{C})
= p(\mathbf{M}_{t-1}\mid \mathbf{M}_t)\;
p(\mathbf{X}_{t-1}\mid \mathbf{X}_t,\mathbf{M}_{t-1},\mathbf{C})\) since the dynamics of $\{\mathbf{M}_t\}_{t=0}^{T}$ are independent of $\{\mathbf{X}_t\}_{t=0}^{T}$.
For the first term \(p(\mathbf{M}_{t-1}\mid \mathbf{M}_t)\), the transition is determined by the visibility schedule \(\{\bar{\alpha}_t\}_{t=1}^{T}\) and is given by $p(\mathbf{M}_{t-1}\mid \mathbf{M}_t)
=\prod_{g=1}^G
\mathrm{Bern}\!\Bigl(M_{t-1}^{(g)};\; M_t^{(g)}+\bigl(1-M_t^{(g)}\bigr)\,\pi_t\Bigr)$, where \(\pi_t=\frac{\bar{\alpha}_{t-1}-\bar{\alpha}_t}{1-\bar{\alpha}_t}\).
This means that if $\mathbf{m}_{t}^{(g)}=1$, then $\mathbf{m}_{t-1}^{(g)}$ must also be $1$; otherwise, \(\mathbf{m}_{t-1}^{(g)}=1\) with probability \(\pi_t\).

We approximate the second term as $p(\mathbf{X}_{t-1}\mid \mathbf{X}_t,\mathbf{X}_0,\mathbf{M}_{t-1},\mathbf{C})$ and parameterize it by $p_{\theta}(\mathbf{X}_{t-1}\mid \mathbf{X}_t,\mathbf{X}_0,\mathbf{M}_{t-1},\mathbf{C})$, where the unknown clean sample is predicted from the current partially observed expression, timestep, and  image condition: \(\hat{\mathbf{x}}_0=f_{\theta}(\mathbf{x}_t,\,t,\,\phi(\mathbf{c}))\).
To ensure consistency between the unmasked components of $\mathbf{x}_{t}$ and $\hat{\mathbf{x}}_{0}$, we apply the calibration: $\tilde{\mathbf{x}}_{0}^{(g)} = \mathbf{m}_{t}^{(g)}\mathbf{x}_{t}^{(g)} + (1-\mathbf{m}_{t}^{(g)})\hat{\mathbf{x}}_{0}^{(g)}$. 
According to Eq.~\ref{eq:fwd}, once $\mathbf{X}_0 = \tilde{\mathbf{x}}_{0}$ and $\mathbf{M}_{t-1}=\mathbf{m}_{t-1}$ are given, the corresponding $\mathbf{x}_{t-1}$ is fully determined \(\delta_{\mathbf{m}_{t-1}\odot \tilde{\mathbf{x}}_0}(\mathbf{X}_{t-1})\). 
Therefore, each reverse transition predicts the masked components of the current partially observed expression $\mathbf{x}_t$ and fills the denoised values into the entries newly activated by $\mathbf{m}_{t-1} - \mathbf{m}_{t}$.

\noindent\textbf{Optimization.}
We optimize $f_{\theta}(\cdot)$ by minimizing the following objective:
\begin{equation}
\mathcal{L}(\theta)
=\mathbb{E}\!\left[
w_t
\left\| \bigl(\mathbf{1}-\mathbf{m}_t\bigr) \odot \bigl(f_\theta(\mathbf{x}_t,t,\phi(\mathbf{c}))-\mathbf{x}_0\bigr)\right\|_2^2
\right],
\label{eq:loss}
\end{equation}
where the expectation is taken over \(t \sim \mathrm{Unif}(\{1,\ldots,T\}) \), \((\mathbf{x}_0,\mathbf{c})\sim p(\mathbf{X},\mathbf{C})\), and $(\mathbf{m}_t, \mathbf{x}_t)$ sampled according to Eq.~\ref{eq:fwd}.
Following masked diffusion models~\cite{zhengreparameterized}, we set the weighting term as \( w_t = \frac{\bar{\alpha}_{t-1} - \bar{\alpha}_t }{1-\bar{\alpha}_t} \). 
In this formulation, $f_{\theta}(\cdot)$ receives partially observed inputs at each step, and the loss is computed only over the masked components, aligning both the input form and supervision pattern with masked autoencoding.
By introducing this stochastic masking process for continuous expression profile, we effectively address the objective mismatch between sc\mbox{-}FM pre-training and generative modeling in ST. 

\noindent\textbf{Inference.} 
Given a histology patch \(\mathbf{c}\), we generate a plausible expression profile \(\mathbf{x}\) by simulating the reverse process.
Specifically, we initialize \(\mathbf{x}_T=\mathbf{0}\) and \(\mathbf{m}_T=\mathbf{0}\).
For each \(t=T,\ldots,1\), we sample \(\mathbf{m}_{t-1}\sim p(\mathbf{M}_{t-1}\mid \mathbf{M}_t=\mathbf{m}_t)\), compute \(\hat{\mathbf{x}}_0=f_\theta(\mathbf{x}_t,\,t,\,\phi(\mathbf{c}))\), set the visibility calibration \(\tilde{\mathbf{x}}_0^{(g)}=\mathbf{m}_t^{(g)}\mathbf{x}_t^{(g)}+(1-\mathbf{m}_t^{(g)})\hat{\mathbf{x}}_0^{(g)}\), and update
\(\mathbf{x}_{t-1}=\mathbf{m}_{t-1}\odot \tilde{\mathbf{x}}_0\).
After \(T\) steps, we obtain \(\mathbf{x}=\mathbf{x}_0 \sim p_{\theta}(\mathbf{X}\mid \mathbf{C}=\mathbf{c})\).
Resampling the mask trajectory yields diverse yet histology-consistent samples.

\subsection{Retrofitting sc-FM with SoftAdaLN}
\label{sec:method:architecture}
To instantiate the denoising network's backbone \(f_\theta(\,\cdot\,)\), we adopt CellFM, a transformer-based sc\mbox{-}FM pre-trained in expression space via masked autoencoding without images~\cite{zeng2025cellfm}. 
We freeze its parameters and retrofit an identity-initialized SoftAdaLN that injects layer-wise context from histology and the timestep while preserving the model’s learned gene--gene dependencies. 
This conditioning mechanism only assumes a token-based, masked-autoencoding backbone and is therefore architecturally compatible with other sc\mbox{-}FMs of this family~\cite{cui2024scscgpt, hao2024large, baek2025scfms}. 
The resulting conditioned transformer serves as our conditional diffusion model's backbone \(f_\theta(\,\cdot\,)\): given $\mathbf{x}_t$, histology $\mathbf{c}$, and the timestep $t$, it outputs $\hat{\mathbf{x}}_0$ for the reverse update in Sec.~\ref{sec:method:masked_diffusion}. 


\noindent\textbf{Token embedding.}
To preserve consistency with CellFM pre-training, we follow its input encoding strategy. For each visible entry in $\mathbf{x}_t$, we apply the value embedding module, while masked entries are assigned a dedicated token ID mapped to a learned embedding, ensuring distinction from true zeros. We then add the corresponding gene-ID embedding to each token, forming a sequence of $G$ tokens in $\mathbb{R}^D$. 

\noindent\textbf{Condition encoding.}
We extract the histology context via a frozen encoder $\phi(\cdot)$, yielding $\mathbf{v}=\phi(\mathbf{c})$. The diffusion timestep $t$ is mapped to an embedding $\mathbf{e}_t$.
These are concatenated and transformed to produce the global condition embedding $\mathbf{c}_t=\varphi_{\text{cond}}([\mathbf{v};\,\mathbf{e}_t])\in\mathbb{R}^{D}$.
This condition embedding modulates all layers of CellFM, providing consistent contextual information across the network.

\noindent\textbf{Condition-driven modulation.}
Each transformer layer in the CellFM backbone comprises two frozen sub-layers: multi-head attention (MHA) and a gated feed-forward module (SGLU). Before each sub-layer, we insert SoftAdaLN, which applies a soft normalization (SoftNorm) followed by identity-initialized affine modulation using the shared condition embedding $\mathbf{c}_t$. For each sub-layer (either MHA or SGLU), denoting its input token embedding as $\mathbf{h}_{\text{in}}\!\in\!\mathbb{R}^D$, SoftNorm is defined as 
\begin{equation}
\mathrm{SoftNorm}(\mathbf{h}_{\text{in}})=(1-\eta)\,\mathbf{h}_{\text{in}}\;+\;\eta\cdot\frac{\mathbf{h}_{\text{in}}-\mu(\mathbf{h}_{\text{in}})}{\sigma(\mathbf{h}_{\text{in}})+\varepsilon},
\label{eq:softnorm}
\end{equation}
with learnable $\eta$ and normalization across $D$ embedding dimensions ($\varepsilon$ prevents division by zero). The full modulation is then given by: 
\begin{equation}
\mathrm{SoftAdaLN}(\mathbf{h}_{\text{in}}\mid\mathbf{c}_t)
=\mathrm{SoftNorm}(\mathbf{h}_{\text{in}})\odot (\mathbf{1}+\mathbf{s}(\mathbf{c}_t)) + \boldsymbol{\kappa}(\mathbf{c}_t),
\label{eq:softadaln}
\end{equation}
where both $\mathbf{s}(\cdot)$ as the scale and $\boldsymbol{\kappa}(\cdot)$ as the shift are $\mathbb{R}^D \rightarrow \mathbb{R}^{D}$ linear layers.

The modulated token embeddings are then fed into the frozen sub-layer, producing the transformed token embeddings.
Let $\mathbf{u}$ denote one such transformed token embedding, it is merged with the residual path via a gated connection and then passed through a frozen pre-trained LayerNorm to produce the final sub-layer output:
\begin{equation}
\mathbf{h}_{\text{out}}
=\mathrm{LN}\!\bigl(\,\boldsymbol{\tau}(\mathbf{c}_t)\odot\mathbf{u}\;+\;\lambda\,\mathbf{h}_{\text{in}}\,\bigr),
\label{eq:gatedres}
\end{equation}
where $\boldsymbol{\tau}(\cdot)\in(0,1]^D$ is a linear layer followed by a sigmoid activation.
We initialize the linear layer with zero weights and a large positive bias so that $\boldsymbol{\tau}(\cdot)$ initially outputs values close to one for all inputs.
$\lambda$ denotes the residual scaling factor inherited from CellFM.
A similar ungated instance of SoftAdaLN is inserted before the decoder to match its input distribution while preserving its pre-trained behavior. 
The decoder then processes the final tokens and outputs $\hat{\mathbf{x}}_0$. This retrofitting enables CellFM to serve as the backbone of our conditional denoising model.

\noindent\textbf{Progressive adaptation without forgetting.}
All modulation components are identity-initialized: $\eta=0$, $\mathbf{s}(\cdot)=\mathbf{0}$, $\boldsymbol{\kappa}(\cdot)=\mathbf{0}$, and $\boldsymbol{\tau}(\cdot)\approx\mathbf{1}$, ensuring that the model initially reproduces its pre-trained behavior.
Transformation $\varphi_{\text{cond}}(\cdot)$ is shared across layers but remains trainable, whereas the linear mappings for $\mathbf{s}(\cdot)$, $\boldsymbol{\kappa}(\cdot)$, and $\boldsymbol{\tau}(\cdot)$ are instantiated per sub-layer to enable layer-wise modulation.
During training, only the modulation parameters $\{\eta,\,\theta_{\varphi},\,\theta_{s},\,\theta_{\kappa},\,\theta_{\tau}\}$ are updated as $\theta$ in Eq.~\ref{eq:loss}, while all other weights, including those of the image encoder and CellFM, remain frozen.  
By freezing the pre-trained CellFM and applying identity initialization, the model initially preserves the existing gene-gene dependencies and then gradually learns to incorporate histology and timestep information through condition-driven modulation, thereby mitigating catastrophic forgetting on limited spatial data. 

\begin{table*}
\centering
\setlength{\tabcolsep}{4pt}
\resizebox{\textwidth}{!}
{
    \begin{tabular}{cl|cccc|cccc|cccc}
    \toprule
    \multicolumn{2}{c|}{\multirow{2}{*}{Methods}} & \multicolumn{4}{c|}{cSCC} & \multicolumn{4}{c|}{Her2ST} & \multicolumn{4}{c}{Kidney}  \\ 
    \cmidrule(lr){3-6} \cmidrule(lr){7-10} \cmidrule(lr){11-14} 
    \multicolumn{2}{c|}{}                        
    & PCC-50 $\uparrow$ & PCC-200 $\uparrow$ & MSE $\downarrow$ & MAE $\downarrow$  
    & PCC-50 $\uparrow$ & PCC-200 $\uparrow$ & MSE $\downarrow$ & MAE $\downarrow$  
    & PCC-50 $\uparrow$ & PCC-200 $\uparrow$ & MSE $\downarrow$ & MAE $\downarrow$ \\ 
    \midrule
    \midrule
    
    \multicolumn{14}{l}{\textbf{Discriminative}} \\
    & ST-Net~\cite{he2020stnet}  
    & \meansd{0.548}{.039} & \meansd{0.448}{.043} & \meansd{1.174}{.109} & \meansd{0.803}{.041}
    & \meansd{0.439}{.030} & \meansd{0.323}{.031} & \meansd{1.132}{.095} & \meansd{0.837}{.039}
    & \meansd{0.327}{.005} & \meansd{0.209}{.005} & \meansd{1.566}{.004} & \meansd{0.982}{.002} \\
    
    & BLEEP~\cite{xie2023BLEEP}  
    & \meansd{0.643}{.009} & \meansd{0.548}{.012} & \meansd{0.925}{.066} & \meansd{0.719}{.035}
    & \meansd{0.520}{.021} & \meansd{0.400}{.018} & \meansd{0.949}{.072} & \meansd{0.754}{.029} 
    & \meansd{0.404}{.011} & \meansd{0.285}{.010} & \meansd{1.493}{.010} & \meansd{0.965}{.003} \\

    & TRIPLEX~\cite{chung2024TRIPLEX}  
    & \meansdul{0.683}{.017} & \meansdul{0.588}{.017} & \meansd{0.904}{.067} & \meansd{0.713}{.025}
    & \meansd{0.536}{.017} & \meansd{0.420}{.018} & \meansd{0.957}{.081} & \meansd{0.768}{.044}
    & \meansdul{0.410}{.031} & \meansdul{0.299}{.025} & \meansdbf{1.315}{.045} & \meansdbf{0.915}{.016} \\
    
    & MERGE~\cite{ganguly2025merge}  
    & \meansd{0.609}{.023} & \meansd{0.510}{.031}  & \meansd{1.082}{.247}  & \meansd{0.788}{.096} 
    & \meansd{0.483}{.051} & \meansd{0.381}{.043}  & \meansd{0.998}{.160}  & \meansd{0.792}{.057} 
    & \meansd{0.242}{.016} & \meansd{0.151}{.015} & \meansd{1.531}{.023} & \meansd{0.969}{.006}   \\

    \multicolumn{14}{l}{\textbf{Generative}} \\
    & Stem~\cite{zhu2025Stem}  
    & \meansd{0.676}{.034} & \meansd{0.577}{.031}  & \meansd{1.267}{.163}  & \meansd{0.817}{.126} 
    & \meansdul{0.559}{.015} & \meansdul{0.433}{.019} & \meansd{0.965}{.144} & \meansd{0.766}{.053} 
    & \meansd{0.388}{.020} & \meansd{0.266}{.014} & \meansd{1.434}{.103} & \meansd{0.940}{.026}   \\

    & STFlow~\cite{huang2025STFlow}  
    & \meansd{0.678}{.013} & \meansd{0.578}{.012}  & \meansdul{0.903}{.096}  & \meansdul{0.706}{.035}
    & \meansd{0.543}{.027} & \meansd{0.425}{.024} & \meansdul{0.929}{.089} & \meansdbf{0.745}{.057}
    & \meansd{0.391}{.004} & \meansd{0.269}{.008} & \meansdul{1.402}{.060} & \meansdul{0.929}{.022}   \\

    & \ours\ \scriptsize{(Ours)}  
    & \meansdbf{0.705}{.006} & \meansdbf{0.613}{.008}  & \meansdbf{0.887}{.081}  & \meansdbf{0.703}{.033} 
    & \meansdbf{0.566}{.008} & \meansdbf{0.446}{.010}  & \meansdbf{0.926}{.047}  & \meansdul{0.757}{.029} 
    & \meansdbf{0.428}{.009} & \meansdbf{0.309}{.010} & \meansd{1.459}{.024} & \meansd{0.964}{.018}   \\
    \bottomrule
    \end{tabular}
}
\caption{
Comparison on cSCC, Her2ST, and Kidney datasets using PCC-50, PCC-200, MSE, and MAE.  
Scores are averaged over test slices and three random seeds, reported as mean~$\pm$~standard deviation. 
Best results are in \textbf{bold}, and second-best are \underline{underlined}. 
}
\label{tab:Quantitative}
\end{table*}
\begin{figure*}[t]
  \centering
   \includegraphics[width=0.98\linewidth]{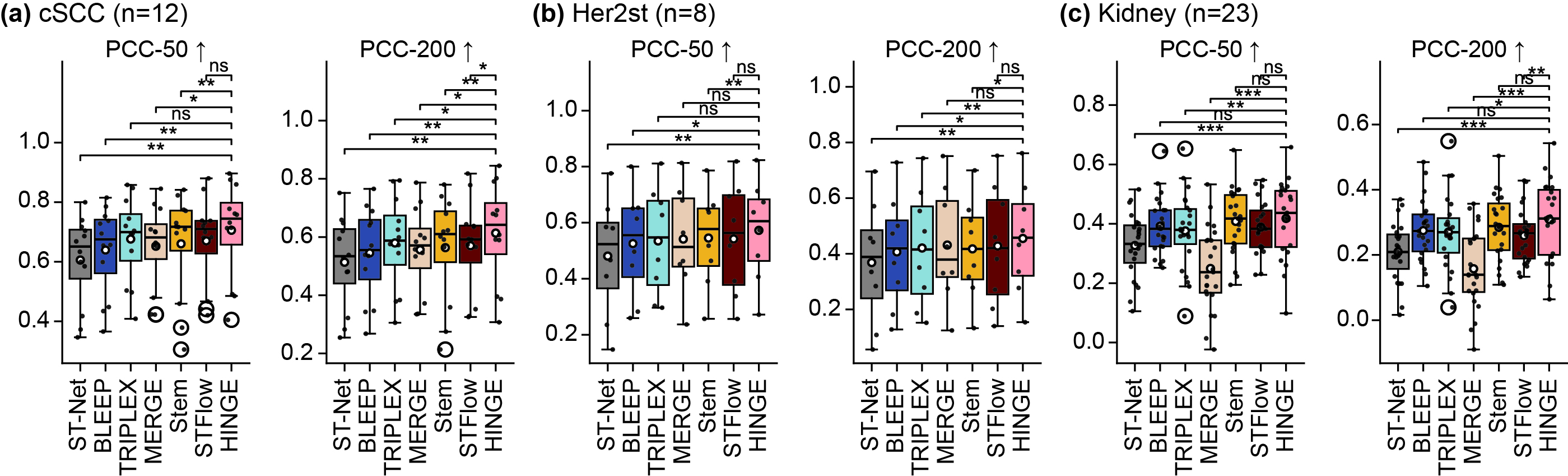}
   \caption{Boxplots of per-slice PCC-50 (left) and PCC-200 (right) with a common random seed for (a) cSCC, (b) Her2ST, and (c) Kidney. Each point represents one test slice. Paired Wilcoxon signed-rank tests assess pairwise differences against baselines.  
   \textbf{*} $p\text{-value}{<}0.05$, \textbf{**} $p\text{-value}{<}0.01$, \textbf{***} $p\text{-value}{<}0.001$, \textbf{ns} not significant. }
   \label{fig3:pvalue}
   \vspace{-0.2in}
\end{figure*}

\subsection{Warm-Start Curriculum}
\label{sec:method:training}
CellFM was pre-trained under masked autoencoding with a low mask ratio (\(\rho\approx 20\%\)).
To respect this regime and stabilize early updates, we begin our training course with a warm-start curriculum.
Specifically, during the initial training phase, the timestep \(t\) is sampled only from a low-mask band, i.e., timesteps with \(\bar{\alpha}_t \ge 1-\rho\), while the scheduler \(\{\bar{\alpha}_t\}\) remains unchanged.
After this warm-start curriculum, we sample over the full range \(t\in\{1,\ldots,T\}\) uniformly.
At all times the masking level is determined solely by the sampled \(t\) rather than the training stage.

\section{Experiments}
\label{sec:Results}
We evaluate \ours\ on three ST datasets to assess its ability to perform histology-conditioned spatial expression generation. 
We first describe the experimental setup, then report quantitative results, visual analyses of marker genes and co-expression patterns, and ablation studies. 

\subsection{Experimental Setup}
\noindent\textbf{Datasets and data preprocessing.}
We conduct experiments on three human ST datasets from different tissues: \textbf{cSCC}~\cite{ji2020SCC} (12 slices from 4 patients), \textbf{Her2ST}~\cite{andersson2020her2st} (36 breast cancer slices from 8 individuals), and \textbf{Kidney}~\cite{lake2023kidney} (23 slices from 22 individuals). Each dataset provides paired H\&E images and spot-level expression matrices. Experimental setup details are summarized in the Appendix. 

Since \ours{} adapts CellFM, which was trained on a fixed 24{,}078-gene vocabulary, we first intersect each ST dataset’s gene list with this vocabulary to ensure compatibility. We then follow the same criteria as in Stem~\cite{zhu2025Stem}, selecting the intersection of genes ranked in the top 300 for both mean expression and variance across training slices, which forms the Highly Mean–High Variance Gene (HMHVG) set used for training, validation, and evaluation. Expression values are log-transformed, consistent with Jaume et al.~\cite{jaume2024hest1k} and Zhu et al.~\cite{zhu2025Stem}, and all inputs and predictions remain in this log scale.

\noindent\textbf{Evaluation protocol.}  
We use a leave-one-slice-out protocol in which each histology slice is held out in turn as the test set. The remaining slices are used for training and validation, with 10\% of the training pool reserved for early stopping. This setup is widely adopted in ST image-to-expression studies.  

Following Stem~\cite{zhu2025Stem}, we evaluate model performance using Pearson correlation coefficients on the 50 and 200 most variable genes within the HMHVG set (denoted as PCC-50 and PCC-200), along with mean squared error (MSE) and mean absolute error (MAE) computed over all genes. All metrics are calculated per test slice and averaged across each dataset. Complete dataset partitions and evaluation protocol details are provided in the Appendix. 

\begin{figure}[t]
  \centering
   \includegraphics[width=0.92\linewidth]{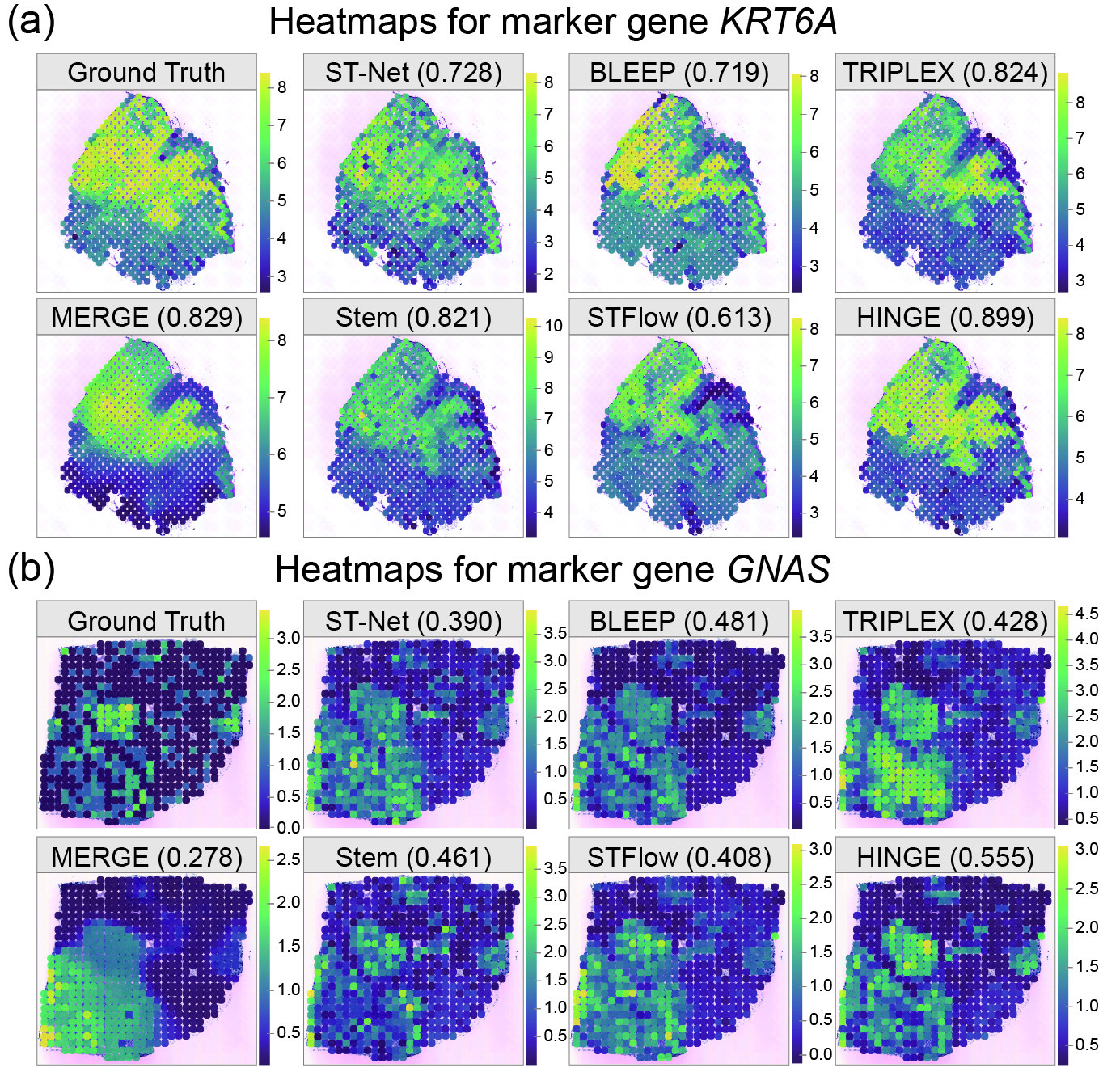}
    \caption{Expression of \textit{KRT6A} on the P2\_ST\_rep3 slice from cSCC (a) and \textit{GNAS} on the H1 slice from Her2ST (b). Each panel shows the ground truth and predictions from representative methods. Both genes are known markers with localized expression.} 
   \label{fig4:maekergene}
   \vspace{-0.2in}
\end{figure}

\noindent\textbf{Baselines.}
We compare \ours{} with six competitive baselines spanning both discriminative and generative paradigms. 
The discriminative group includes four models that directly predict gene expression from histology.  
\textbf{ST-Net}~\cite{he2020stnet} is a CNN-based regressor trained to map histology images to gene profiles.  
\textbf{BLEEP}~\cite{xie2023BLEEP} uses bi-modal contrastive learning to align histology patches with expression references and performs nearest-neighbor imputation.  
\textbf{TRIPLEX}~\cite{chung2024TRIPLEX} extracts multi-scale image representations across tissue hierarchies to inform regression.  
\textbf{MERGE}~\cite{ganguly2025merge} constructs a graph over image patches and propagates features through a hierarchical graph neural network. 

The generative group includes two recent models that synthesize expression patterns from histology.   
\textbf{Stem}~\cite{zhu2025Stem} applies conditional diffusion, where histology features guide the generative trajectory.   
\textbf{STFlow}~\cite{huang2025STFlow} employs flow matching to learn a spatial transport map from histology to expression. 
All baselines are trained and evaluated on the same HMHVG set, with log-transformed expression values.

\subsection{Quantitative Results}
We evaluate the effectiveness of adapting a sc\mbox{-}FM for histology-conditioned spatial expression generation by comparing \ours\ against discriminative and generative baselines.  
Table~\ref{tab:Quantitative} summarizes slice-level performance on three ST datasets in terms of PCC-50, PCC-200, MSE, and MAE. All scores are averaged over test slices and three random seeds (mean $\pm$ standard deviation). 

On \textbf{cSCC}, \ours{} achieves the best performance across all metrics. 
Relative to the strongest generative baseline STFlow, it improves PCC-50 from 0.678 to 0.705 (about 4\%) and PCC-200 from 0.578 to 0.613 (about 6\%), and it also surpasses the strongest discriminative model TRIPLEX by roughly 3.2\% and 4.3\% in PCC-50 and PCC-200, respectively. 
On \textbf{Her2ST}, \ours{} again attains the highest PCC-50 and PCC-200 (0.566 and 0.446), corresponding to gains of about 1.3\% and 3.0\% over the best baseline (Stem), and also achieves the lowest MSE and a competitive MAE. 
On \textbf{Kidney}, \ours{} yields the strongest correlation scores, improving PCC-50 from 0.410 to 0.428 (about 4.4\%) and PCC-200 from 0.299 to 0.309 (about 3.3\%) compared to TRIPLEX, while its MSE and MAE are higher than those of the best discriminative baseline but remain close to those of the generative baselines (e.g., MSE 1.459 vs.\ 1.402 for STFlow), indicating that on this dataset its advantage is mainly reflected in correlation-based metrics, with absolute errors comparable to other generative methods.  

To assess robustness across slices, we visualize per-slice PCC distributions with a common random seed in Fig.~\ref{fig3:pvalue}. 
\ours{} achieves the highest median PCC across all datasets. 
Paired Wilcoxon signed-rank tests yield significant differences ($p\text{-value} < 0.05$) on multiple datasets, particularly cSCC. 
\ours{} also surpasses all baselines on the majority of slices in terms of PCC, indicating that the observed improvements reflect consistent trends rather than being driven by a few outlier slices. We also report gene-wise structural similarity (SSIM) in the Appendix. 

\begin{figure}[t]
  \centering
   \includegraphics[width=1\linewidth]{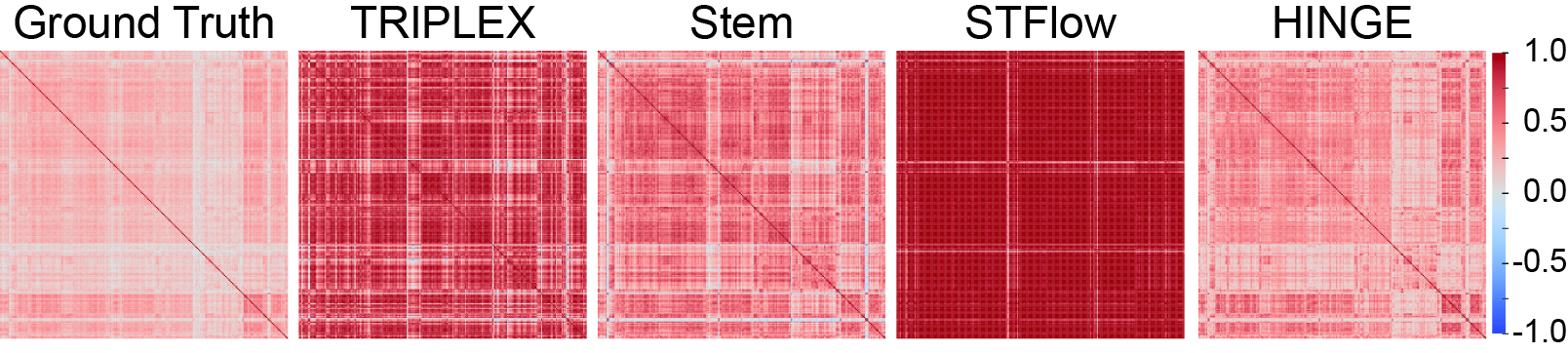}
    \caption{Gene–gene correlation matrices computed on the DKD Kidney slice 31-10042. Each matrix is derived from the predicted expression values of the HMHVG gene set. The ground truth and representative method outputs are shown. }
   \label{fig5:corrmatrix}
   \vspace{-0.2in}
\end{figure}

\subsection{Marker Expression and Gene Correlation} 
We next assess whether the predicted expression aligns with known biological patterns by examining the output on marker genes and gene–gene relationships. 
Following prior histology-conditioned generative models such as Stem~\cite{zhu2025Stem} and STFlow~\cite{huang2025STFlow}, we treat these structure-level analyses as complementary to spot-wise metrics. 

To evaluate marker expression fidelity, we visualize the predicted expression values of \textit{KRT6A} on the P2\_ST\_rep3 slice (cSCC)~\cite{ji2020SCC} and \textit{GNAS} on the H1 slice (Her2ST)~\cite{andersson2020her2st}, shown in Fig.~\ref{fig4:maekergene}. 
Both genes are established markers with localized expression: \textit{KRT6A} is associated with squamous cell carcinoma, while \textit{GNAS} is implicated in breast cancer signaling. 
\ours{} more accurately captures the high-expression regions seen in the ground truth, preserving the spatial contrast and avoiding the oversmoothing seen in several baselines. 
These results suggest that our model produces biologically coherent spatial expression patterns for tissue-specific marker genes. 

To examine gene–gene co-expression, we plot correlation matrices for a DKD Kidney slice (31-10042) in Fig.~\ref{fig5:corrmatrix}. \ours{} better matches the ground-truth correlation structure, preserving both strong and weak co-expression, whereas ST-only baselines tend to blur these patterns. 
This suggests that \ours{} preserves gene dependencies encoded by the sc\mbox{-}FM while using histology to adapt them to the spatial context. 
More visualizations are in the Appendix. 

\begin{table}[t]
\centering
\renewcommand{\arraystretch}{0.8}    
\resizebox{\columnwidth}{!}{
\begin{tabular}{l|cccc}
\toprule
Variant & PCC-50 $\uparrow$ & PCC-200 $\uparrow$ & MSE $\downarrow$ & MAE $\downarrow$ \\ 
\midrule
Scratch  & 0.7425  & 0.6395 & 2.6175 & 1.1834  \\ 
Decoder-Tune        & 0.8518 & 0.7699 & 1.3275 & 0.8999   \\ 
Backbone-LoRA         & 0.8457 & 0.7618 & 1.2706 & 0.8800    \\ 
HINGE  & \textbf{0.8755} & \textbf{0.8021} & \textbf{1.0096} & \textbf{0.7793}  \\ 
\bottomrule
\end{tabular}
}
\caption{Effect of reusing a pre-trained sc\mbox{-}FM under different adaptation schemes.} 
\label{tab:ablation-pretrain}
\end{table}

\begin{table}[t]
\centering
\renewcommand{\arraystretch}{0.8}    
\resizebox{\columnwidth}{!}{
\begin{tabular}{l|cccc}
\toprule
Variant & PCC-50 $\uparrow$ & PCC-200 $\uparrow$ & MSE $\downarrow$ & MAE $\downarrow$ \\
\midrule
Gauss-Diff    & 0.7738 & 0.6415 & 1.7249 & 0.9855 \\
Mask-Diff (NoCurr) & 0.8691 & 0.7932 & 1.1619 & 0.8483 \\
Mask-Diff (RandMask)  & 0.8752 & 0.7996 & 1.0208 & 0.7869 \\
Mask-Diff (HINGE)   & \textbf{0.8755} & \textbf{0.8021} & \textbf{1.0096} & \textbf{0.7793}  \\ 
\bottomrule
\end{tabular}}
\caption{Comparison of Gaussian diffusion, masked diffusion variants, and the full HINGE objective. }
\label{tab:ablation-objective}
\vspace{-0.15in}
\end{table}

\subsection{Ablation Studies}
\label{sec:ablation}
We ablate four main design choices in \ours{}: pre-trained sc\mbox{-}FM initialization and fine-tuning, generative objective, histology conditioning within the backbone, and histology encoder. Unless noted otherwise, experiments use cSCC (P2\_ST\_rep3), with more results in the Appendix.

\noindent\textbf\noindent\textbf{Whether pre-trained sc\mbox{-}FM helps and which fine-tuning strategy is better?} 
Table~\ref{tab:ablation-pretrain} compares how different update schemes use pre-training in \ours{}. All variants share the same \ours{} architecture with trainable SoftAdaLN modulators. \textbf{Scratch} learns all parameters from random initialization, instead of inheriting from a pre-trained CellFM.
\textbf{Decoder-Tune} finetunes the pre-trained decoder. 
\textbf{Backbone-LoRA} augments each attention and feed-forward sublayer with LoRA adapters, enabling low-rank backbone updates. 
\textbf{HINGE} keeps all pre-trained weights frozen and optimizes only the conditioned modulators. 
Scratch performs worst, while \textbf{HINGE} achieves the best scores, indicating that a pre-trained sc\mbox{-}FM is helpful for ST and that, under limited ST supervision, keeping the pre-trained weights frozen is more effective and helps mitigate catastrophic forgetting. 

\noindent\textbf{How do the generative objective and corruption process affect transfer from a pre-trained sc\mbox{-}FM?} 
Table~\ref{tab:ablation-objective} compares different generative objectives and corruption mechanisms under the same \ours{} architecture. 
\textbf{Gauss-Diff} replaces masked diffusion with DDPM-style Gaussian diffusion over all expression dimensions and yields the weakest results, indicating that corrupting all input dimensions with Gaussian noise introduces an input–supervision mismatch that can hinder transfer of sc-FM knowledge. 
\textbf{Mask-Diff (NoCurr)} uses masked diffusion without the warm-start curriculum and shows a slight drop relative to \textbf{Mask-Diff (HINGE)}, suggesting that emphasizing low-mask steps early stabilizes optimization. 
\textbf{Mask-Diff (RandMask)} fills masked coordinates with random rather than zero values but still matches Mask-Diff (HINGE), consistent with our input encoding, which maps masked entries to a mask-token embedding distinct from true zeros and makes the expression-space placeholder value largely irrelevant. 
We further analyze masked diffusion sampling via the denoising trajectory, the inference-step budget, and sensitivity to $T$ and the visibility schedule $\zeta$; see Appendix.  

\begin{table}[t]
\centering
\renewcommand{\arraystretch}{0.8}    
\resizebox{\columnwidth}{!}{
\begin{tabular}{l|cccc}
\toprule
Variant & PCC-50 $\uparrow$ & PCC-200 $\uparrow$ & MSE $\downarrow$ & MAE $\downarrow$ \\ 
\midrule
Hist-Affine-LN & 0.8298 & 0.7690 & 1.7371 & 0.9809     \\ 
SoftAdaLN (NoSoftNorm)  &  0.8658 & 0.7874 & 1.2317 & 0.8699 \\
SoftAdaLN (NoIdInit)   & 0.8631 & 0.7722 & 1.2071 & 0.8600  \\  
SoftAdaLN (Full)   & \textbf{0.8755} & \textbf{0.8021} & \textbf{1.0096} & \textbf{0.7793}  \\  
\bottomrule
\end{tabular}
}
\caption{Comparison of alternative conditioning mechanisms.} 
\label{tab:ablation-conditioning}
\end{table}

\begin{table}[t]
\centering
\renewcommand{\arraystretch}{0.8}    
\resizebox{\columnwidth}{!}{
\begin{tabular}{l|cccc}
\toprule
Variant & PCC-50 $\uparrow$ & PCC-200 $\uparrow$ & MSE $\downarrow$ & MAE $\downarrow$ \\
\midrule 
UNI           & 0.8625 & 0.7871 & 1.1512 & 0.8406 \\
CONCH         & 0.8613 & 0.7783 & 1.3316 & 0.9029 \\
UNI + CONCH   & \textbf{0.8755} & \textbf{0.8021} & \textbf{1.0096} & \textbf{0.7793}  \\ 
\bottomrule
\end{tabular}}
\caption{Impact of different histology encoders. }
\label{tab:ablation-histenc}
\vspace{-0.15in}
\end{table}

\noindent\textbf{How should histology be injected into the frozen backbone?} 
Table~\ref{tab:ablation-conditioning} compares alternative ways to inject histology into the frozen backbone. 
\textbf{Hist-Affine-LN} replaces the original post-layer normalization with a histology-conditioned affine layer on the normalization scale and shift; this aggressive modification yields the largest degradation, suggesting that directly overwriting normalization statistics can disrupt pre-trained representations. \textbf{SoftAdaLN (NoSoftNorm)} removes the SoftNorm component, while \textbf{SoftAdaLN (NoIdInit)} keeps the full structure but drops identity initialization; both variants fall short of \textbf{SoftAdaLN (Full)}, indicating that SoftNorm and identity initialization stabilize conditioning.

\noindent\textbf{Which histology encoder provides the most effective conditioning?} 
Finally, Table~\ref{tab:ablation-histenc} compares histology encoders used to condition \ours{}. We evaluate \textbf{UNI}~\cite{chen2024UNI}, \textbf{CONCH}~\cite{lu2024CONCH}, and their concatenation \textbf{UNI+CONCH}. UNI and CONCH achieve similar performance, whereas UNI+CONCH performs best, suggesting that global context from UNI and attention-based features from CONCH provide complementary histology cues for conditioning. Further variants are reported in the Appendix.

\section{Conclusion}
\label{sec:Conclusion}
We present \ours{}, a novel framework that adapts pre-trained expression-only single-cell foundation models to histology-conditioned spatial expression generation.
\ours{} combines identity-initialized modulation via SoftAdaLN with a masked diffusion objective and a simple timestep sampling scheme, enabling stable knowledge transfer from masked autoencoding sc\mbox{-}FMs.
Experiments on three ST datasets show that \ours{} outperforms baselines in accuracy, spatial coherence, and co-expression fidelity. 
Although instantiated on CellFM in this work, our conditioning design is architecture-agnostic and can, in principle, be applied to other sc\mbox{-}FMs (e.g., scGPT), offering a general pathway for incorporating sc\mbox{-}FMs into histology-based tissue modeling. 

\section*{Acknowledgements}
This work was supported by the Guangdong S\&T \mbox{Program} (Grant No. 2024B0101040005), the \mbox{National} Natural Science Foundation of China (Grant Nos. 62262069, 62302537, 62402071), and the Guangdong \mbox{Basic} and Applied Basic Research Foundation (Grant No. 2025A1515012856). 

\section*{Appendices}
Generating spatial gene expression profiles from histology images helps mitigate the high cost and limited accessibility of spatial transcriptomics (ST). This approach can improve tissue analysis—including spatial domain identification and biomarker discovery—and provide insights into cellular interactions within tissues, ultimately facilitating the identification of disease biomarkers and advancing clinical applications \cite{min2025spamask, fang2025spacross, niu2025spabatch, xue2025inferring}. 

\appendix
\renewcommand{\thesection}{\Alph{section}}
\setcounter{figure}{0}
\setcounter{table}{0}
\renewcommand{\thefigure}{S\arabic{figure}}
\renewcommand{\thetable}{S\arabic{table}}

\section{Experimental Setup Details}
This section provides an expanded description of the experimental setup, complementing the settings outlined in the main paper. 

\subsection{Datasets Description}

We employed three publicly available spatial transcriptomics (ST) datasets covering distinct tissue types, disease contexts, and experimental platforms (as summarized in Table~\ref{tab:dataset_summary}).

(1) \textbf{cSCC (ST, Cutaneous Squamous Cell Carcinoma).}  
The cSCC dataset~\cite{ji2020SCC} consists of formalin-fixed paraffin-embedded (FFPE) cutaneous squamous cell carcinoma samples from four patients, profiled with the Spatial Transcriptomics platform using a spot grid with 110~$\mu$m center-to-center spacing and 150~$\mu$m spot diameter. The slices exhibit highly heterogeneous tumor microenvironments, including keratinized tumor nests, stromal regions, and immune cell infiltrates. Although FFPE processing typically leads to reduced RNA integrity, this dataset offers a realistic and diverse benchmark for assessing model robustness under degraded molecular quality.

(2) \textbf{Her2ST (ST, HER2$^{+}$ Breast Cancer).}  
The Her2ST dataset~\cite{andersson2020her2st} comprises spatial transcriptomics measurements of HER2-positive invasive ductal carcinoma (IDC) from eight patients. Each slice was captured using the original ST protocol on fresh-frozen breast tissue, with a spot diameter of 100~$\mu$m and an inter-spot distance of 200~$\mu$m. In total, 36 slices were included, covering tumor core and peritumoral regions. The dataset provides high-quality histology images aligned with corresponding spot-level gene expression matrices ($\approx$15k detected genes per slice).

(3) \textbf{Human Kidney (Visium ST).}  
The Kidney dataset~\cite{lake2023kidney} represents a large-scale Visium Spatial Gene Expression collection containing 23 patient samples spanning both healthy and diseased conditions (Diabetic Kidney Disease and Acute Kidney Injury). All slices were obtained from fresh-frozen human Kidney tissue with a 55~$\mu$m spot diameter and 100~$\mu$m inter-spot distance. Each slice contains between 0.3k–4k spots with over 33k expressed genes, capturing both cortical and medullary regions at high molecular depth. This dataset provides extensive biological and technical variability, serving as the primary benchmark for assessing generalization across tissues and disease states.

\subsection{Dataset Partitioning}
Each histology slice is held out in turn as the test set, while the remaining slices are used for model training and validation. From the training pool, 10\% of samples are randomly reserved as a validation subset to monitor model performance and trigger early stopping. This design ensures that model assessment is always performed on unseen tissue slices, thereby preventing data leakage and providing a reliable measure of generalization across tissue slices.

Specially for the Her2ST dataset, which provides 3--6 serial slices per patient across eight patients, we designate the first slice from each patient (A1, B1, C1, D1, E1, F1, G1, H1) as a fixed pool of evaluation candidates. In each evaluation fold, one of these eight slices is held out as the test slice, while all remaining slices, including other slices from the same patient, are used for training, with 10\% of the training samples reserved for validation. This fixed pool with one representative slice per patient keeps the difficulty of different folds comparable and guarantees that every patient is evaluated on its own test slice.

For the cSCC and Kidney datasets, we adopt the same slice-wise training–testing scheme. In each fold, one histology slice is held out for testing, and the remaining slices form the training pool, from which 10\% of samples are reserved for validation. This unified evaluation protocol provides a consistent and fair basis for assessing model generalization across heterogeneous tissues, disease states, and ST platforms.

\subsection{Genes Selection}
\begin{figure*}[t]
    \centering
    \includegraphics[width=\textwidth]{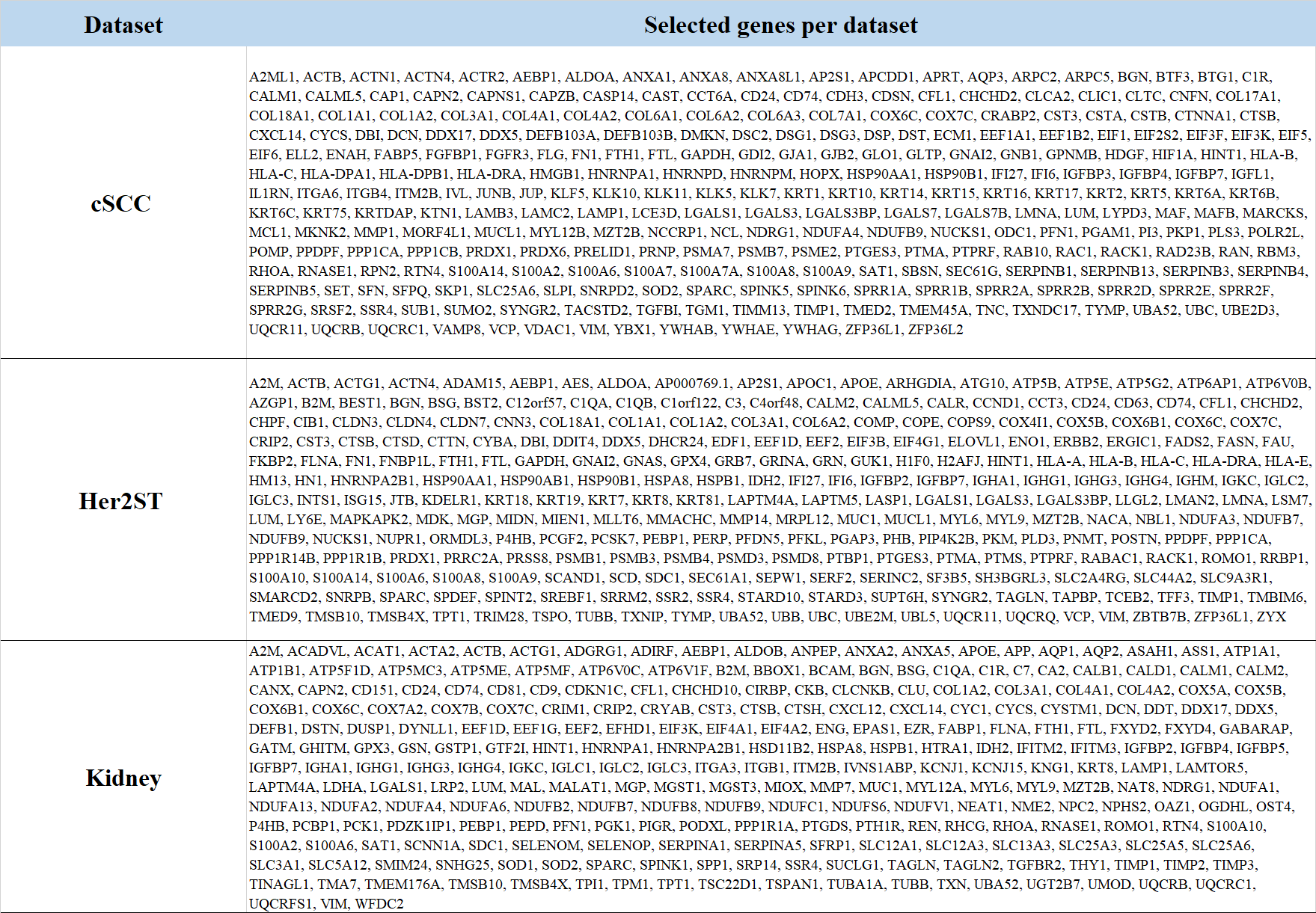}
    \caption{
        Selected genes used across different datasets.
    }
    \label{fig:genes_select}
\end{figure*}

Since \ours{} adapts CellFM, which was trained on a fixed 24{,}078-gene vocabulary, we first intersect each ST dataset’s gene list with this vocabulary to ensure compatibility. We then follow the same criteria as in Stem~\cite{zhu2025Stem}, selecting the intersection of genes ranked in the top 300 for both mean expression and variance across training slices, which forms the Highly Mean–High Variance Gene (HMHVG) set used for training, validation, and evaluation. The selected genes are summarized in Fig.~\ref{fig:genes_select}. 

\begin{table*}[t]
\centering
\caption{Summary of spatial transcriptomics datasets aggregated by patient. Each patient entry reports spot- and gene-level ranges across multiple slices, together with the corresponding platform.}
\resizebox{\textwidth}{!}{
\begin{tabular}{cccccccccccc}
\toprule
\textbf{Dataset} & \textbf{Platform} & \textbf{Patient} & \textbf{Tissue} & \textbf{Condition} & \textbf{Samples} & \textbf{Inter-spot Dist ($\mu$m)} & \textbf{Spot Diameter ($\mu$m)} & \textbf{Spots under Tissue} & \textbf{Genes per slice} & \textbf{Preservation} & \textbf{Reference} \\
\midrule
Her2ST & ST & Patient A & Breast & Cancer & 6 & 200 & 100 & 325--360 & 15,045--15,645 & Fresh Frozen & \href{https://pubmed.ncbi.nlm.nih.gov/34650042/}{PMID: 34650042} \\
 & ST & Patient B & Breast & Cancer & 6 & 200 & 100 & 270--295 & 15,109--15,387 & Fresh Frozen & same as above \\
 & ST & Patient C & Breast & Cancer & 6 & 200 & 100 & 176--187 & 15,557--15,842 & Fresh Frozen & same as above \\
 & ST & Patient D & Breast & Cancer & 6 & 200 & 100 & 301--315 & 15,396--15,666 & Fresh Frozen & same as above \\
 & ST & Patient E & Breast & Cancer & 3 & 200 & 100 & 570--587 & 15,097--15,701 & Fresh Frozen & same as above \\
 & ST & Patient F & Breast & Cancer & 3 & 200 & 100 & 691--712 & 14,861--15,067 & Fresh Frozen & same as above \\
 & ST & Patient G & Breast & Cancer & 3 & 200 & 100 & 441--467 & 14,992--15,258 & Fresh Frozen & same as above \\
 & ST & Patient H & Breast & Cancer & 3 & 200 & 100 & 510--613 & 14,873--15,029 & Fresh Frozen & same as above \\
\midrule
cSCC & ST & Patient 2 & Skin & Cancer & 3 & 110 & 150 & 638--666 & 17,138--17,883 & FFPE & \href{https://www.ncbi.nlm.nih.gov/pmc/articles/PMC7391009/}{PMID: 7391009} \\
 & ST & Patient 5 & Skin & Cancer & 3 & 110 & 150 & 521--590 & 16,959--17,689 & FFPE & same as above \\
 & ST & Patient 9 & Skin & Cancer & 3 & 110 & 150 & 1071--1182 & 17,823--19,314 & FFPE & same as above \\
 & ST & Patient 10 & Skin & Cancer & 3 & 110 & 150 & 462--621 & 15,383--17,047 & FFPE & same as above \\
\midrule
Kidney & Visium ST & Patient 1 & Kidney & Healthy & 1 & 100 & 55 & 3007 & 33538 & Fresh Frozen & \href{https://www.ncbi.nlm.nih.gov/pmc/articles/PMC10356613/}{PMID: 10356613} \\
 & Visium ST & Patient 2 & Kidney & Healthy & 1 & 100 & 55 & 3627 & 36601 & Fresh Frozen & same as above \\
 & Visium ST & Patient 3 & Kidney & Healthy & 1 & 100 & 55 & 4166 & 36601 & Fresh Frozen & same as above \\
 & Visium ST & Patient 4 & Kidney & Healthy & 1 & 100 & 55 & 2627 & 36601 & Fresh Frozen & same as above \\
 & Visium ST & Patient 5 & Kidney & Healthy & 1 & 100 & 55 & 956 & 36601 & Fresh Frozen & same as above \\
 & Visium ST & Patient 6 & Kidney & Healthy & 1 & 100 & 55 & 1034 & 36601 & Fresh Frozen & same as above \\
\midrule
 & Visium ST & Patient 7 & Kidney & Diseased & 1 & 100 & 55 & 1322 & 36601 & Fresh Frozen & same as above \\
 & Visium ST & Patient 8 & Kidney & Diseased & 1 & 100 & 55 & 673 & 36601 & Fresh Frozen & same as above \\
 & Visium ST & Patient 9 & Kidney & Diseased & 1 & 100 & 55 & 673 & 36601 & Fresh Frozen & same as above \\
 & Visium ST & Patient 10 & Kidney & Diseased & 1 & 100 & 55 & 560 & 36601 & Fresh Frozen & same as above \\
 & Visium ST & Patient 11 & Kidney & Diseased & 1 & 100 & 55 & 534 & 36601 & Fresh Frozen & same as above \\
 & Visium ST & Patient 12 & Kidney & Diseased & 1 & 100 & 55 & 453 & 36601 & Fresh Frozen & same as above \\
 & Visium ST & Patient 13 & Kidney & Diseased & 2 & 100 & 55 & 461-904 & 36601 & Fresh Frozen & same as above \\
 & Visium ST & Patient 14 & Kidney & Diseased & 1 & 100 & 55 & 601 & 36601 & Fresh Frozen & same as above \\
 & Visium ST & Patient 15 & Kidney & Diseased & 1 & 100 & 55 & 787 & 36601 & Fresh Frozen & same as above \\
 & Visium ST & Patient 16 & Kidney & Diseased & 1 & 100 & 55 & 407 & 36601 & Fresh Frozen & same as above \\
 & Visium ST & Patient 17 & Kidney & Diseased & 1 & 100 & 55 & 317 & 36601 & Fresh Frozen & same as above \\
 & Visium ST & Patient 18 & Kidney & Diseased & 1 & 100 & 55 & 645 & 36601 & Fresh Frozen & same as above \\
 & Visium ST & Patient 19 & Kidney & Diseased & 1 & 100 & 55 & 673 & 36601 & Fresh Frozen & same as above \\
 & Visium ST & Patient 20 & Kidney & Diseased & 1 & 100 & 55 & 640 & 36601 & Fresh Frozen & same as above \\
 & Visium ST & Patient 21 & Kidney & Diseased & 1 & 100 & 55 & 507 & 36601 & Fresh Frozen & same as above \\
 & Visium ST & Patient 22 & Kidney & Diseased & 1 & 100 & 55 & 370 & 36601 & Fresh Frozen & same as above \\
\bottomrule
\end{tabular}}
\label{tab:dataset_summary}
\end{table*}

\subsection{Histology Feature Extraction}
Our pipeline operates on spot-level spatial transcriptomics data, where each spot corresponds to a spatial location on an H\&E-stained tissue slide with paired gene expression and histology signals.

Concretely, each spot is associated with an RGB image patch centered at its spatial coordinates $(x_c, y_c)$ on the registered whole-slide image (WSI). Patch extraction proceeds as follows:
\begin{enumerate}
    \item \textbf{Coordinate-based cropping}: Given $(x_c, y_c)$, we extract a square region centered at the spot on the H\&E WSI.
    \item \textbf{Pixel normalization}: Raw pixel intensities in $[0, 255]$ are linearly rescaled to $[0, 1]$ before feeding patches into the encoders.
\end{enumerate}

We adopt a dual-encoder framework to jointly capture complementary visual information from histopathology images. Each normalized image patch is independently encoded by two pre-trained vision backbones:

\begin{itemize}
\item \textbf{UNI}~\cite{chen2024UNI}: A pathology-domain vision transformer pre-trained on large-scale histopathology datasets. It produces embeddings $\mathbf{h}_{\text{uni}} \in \mathbb{R}^{1024}$ that emphasize detailed morphology, including cellular topology, nuclear texture, and tissue microarchitecture.

\item \textbf{CONCH}~\cite{lu2024CONCH}: A contrastive vision--language foundation model trained to align histopathology images with expert pathology reports. It outputs embeddings $\mathbf{h}_{\text{conch}} \in \mathbb{R}^{512}$ that capture high-level semantic tissue context and pathological attributes.
\end{itemize}

The two encoders provide complementary representations—UNI focuses on structural and morphological fidelity, whereas CONCH encodes semantic and contextual information from cross-modal supervision. Their outputs are concatenated to form a unified visual representation for each spot:
\begin{equation}
\mathbf{v}= \phi(\mathbf{c}) = [\mathbf{h}_{\text{uni}}; \mathbf{h}_{\text{conch}}] \in \mathbb{R}^{1536},
\label{eq:image_embedding}
\end{equation}
which serves as the histology feature conditioning the spatial expression generation network.

\subsection{Baselines}
\textbf{ST-Net}~\cite{he2020stnet} integrates spatial transcriptomics and histology through a deep convolutional network to predict gene expression directly from H\&E-stained images. It employs a DenseNet-121 backbone~\cite{huang2017densely} pre-trained on ImageNet, with a fully connected regression head for gene-level prediction. Each 224×224 patch centered on a spatial spot serves as input. Our implementation retains the original network configuration and normalization strategy to ensure faithful reproduction of the published framework.

\textbf{BLEEP}~\cite{xie2023BLEEP} proposes a bi-modal embedding framework for spatial gene expression prediction from H\&E-stained histology images. It aligns image and expression modalities through contrastive learning to construct a shared embedding space, using a ResNet-50 encoder for images and an MLP for expression features. Gene expression is inferred via query–reference imputation based on the proximity of embeddings in the joint space. We adopt the same dual-encoder architecture and contrastive alignment scheme as described in the original work.

\textbf{TRIPLEX}~\cite{chung2024TRIPLEX} introduces a multi-resolution deep learning framework for predicting spatial gene expression from whole-slide histology images. The model captures complementary information at three hierarchical levels—the target spot, its local neighborhood, and the global tissue context—using independent ResNet-based encoders followed by a transformer-based fusion layer. These representations are integrated through an efficient fusion mechanism to jointly model fine-grained morphology and global organization. In our reimplementation, we preserve the multi-resolution design and fusion strategy to maintain methodological fidelity to the original model.

\textbf{MERGE}~\cite{ganguly2025merge} introduces a graph-based framework for spatial gene expression prediction from whole-slide histology images. It constructs a multi-faceted hierarchical graph where nodes represent tissue patches, and edges capture both spatial and morphological relationships. Using a ResNet18 encoder to extract patch features, MERGE employs a Graph Attention Network (GAT) to jointly model short- and long-range dependencies across the tissue. The hierarchical graph integrates intra-cluster and inter-cluster connections, enabling efficient information propagation between morphologically similar but spatially distant regions. We follow the original multi-faceted graph design and SPCS-based gene smoothing to reproduce its morphology-aware prediction behavior

\textbf{Stem}~\cite{zhu2025Stem} introduces a diffusion-based generative framework for predicting spatially resolved gene expression from H\&E-stained histology images. Instead of treating prediction as deterministic regression, Stem models the conditional distribution of gene expression given image features, enabling one-to-many mappings that capture biological heterogeneity. The model leverages pretrained pathology foundation encoders (UNI~\cite{chen2024UNI}, CONCH~\cite{lu2024CONCH}) to derive image embeddings and conditions a DiT-based diffusion network for expression generation. This design allows Stem to generate biologically diverse yet accurate predictions across spatial locations. In our implementation, we maintain the same conditional diffusion formulation and foundation-model conditioning strategy as described in the original paper.

\textbf{STFlow}~\cite{huang2025STFlow} formulates spatial gene expression prediction as a generative modeling problem via whole-slide flow matching. Instead of independent spot-level regression, it models the joint distribution of gene expressions across all spatial locations, capturing cell–cell interactions and global dependencies. The model employs an $E(2)$-invariant Transformer denoiser with local spatial attention and leverages pretrained pathology foundation encoders for feature extraction. We adopt the same flow matching formulation and spatial attention architecture as described in the original work to ensure methodological consistency across baselines.

\begin{table}[t]
\centering
\resizebox{\columnwidth}{!}{
\begin{tabular}{l|cccc}
\toprule
Variant & PCC-50 $\uparrow$ & PCC-200 $\uparrow$ & MSE $\downarrow$ & MAE $\downarrow$ \\ 
\midrule
Scratch  & 0.2511  & 0.1804 & 1.9176 & 1.1215  \\ 
Decoder-Tune        & 0.4726 & \textbf{0.3374} & 0.9814 & 0.7716   \\ 
Backbone-LoRA         & 0.4599 & 0.3163 & 0.9975 & 0.7774    \\ 
HINGE  & \textbf{0.4801} & {0.3355} & \textbf{0.9481} & \textbf{0.7638}  \\ 
\midrule
\midrule
Gauss-Diff    & 0.3437 & 0.2084 & 1.8403 & 1.0447 \\
Mask-Diff (NoCurr) & 0.4702 & 0.3203 & 0.9641 & 0.7691 \\
Mask-Diff (RandMask)  & \textbf{0.4889} & \textbf{0.3427} & 0.9500 & 0.7729 \\
Mask-Diff (HINGE)   & {0.4801} & {0.3355} & \textbf{0.9481} & \textbf{0.7638}  \\ 
\midrule
\midrule
Hist-Affine-LN & 0.2892 & 0.2187 & 1.9070 & 1.1176     \\ 
SoftAdaLN (NoSoftNorm)  &  0.3707 & 0.2432 & 1.3417 & 0.9050 \\
SoftAdaLN (NoIdInit)   & 0.4008 & 0.2873 & 1.2219 & 0.8592  \\  
SoftAdaLN (Full)   & \textbf{0.4801} & \textbf{0.3355} & \textbf{0.9481} & \textbf{0.7638}  \\ 
\midrule
\midrule
ResNet-50     & 0.4348 & 0.2956 & 0.9945 & 0.8001 \\
UNI           & 0.4668 & 0.3208 & 0.9530 & 0.7667 \\
CONCH         & 0.4091 & 0.2630 & 1.0895 & 0.8133 \\
UNI + CONCH   & \textbf{0.4801} & \textbf{0.3355} & \textbf{0.9481} & \textbf{0.7638}  \\ 
\bottomrule
\end{tabular}
}
\caption{Ablations on Her2ST. Component-wise analysis of HINGE variants on the Her2ST (A1) dataset.}
\label{tab:ablation-her2st}
\end{table}

\begin{table}[t]
\centering
\resizebox{\columnwidth}{!}{
\begin{tabular}{l|cccc}
\toprule
Variant & PCC-50 $\uparrow$ & PCC-200 $\uparrow$ & MSE $\downarrow$ & MAE $\downarrow$ \\ 
\midrule
Scratch  & 0.2517  & 0.2072 & 1.7576 & 1.1192  \\ 
Decoder-Tune        & 0.4702 & 0.3627 & 0.9066 & 0.7552   \\ 
Backbone-LoRA         & 0.4558 & 0.3271 & 1.0054 & 0.7984    \\ 
HINGE  & \textbf{0.4871} & \textbf{0.3815} & \textbf{0.8956} & \textbf{0.7467}  \\ 
\midrule
\midrule
Gauss-Diff    & 0.2592 & 0.1673 & 1.7559 & 1.0054 \\
Mask-Diff (NoCurr) & 0.4725 & 0.3735 & 0.9086 & 0.7591 \\
Mask-Diff (RandMask)  & 0.4707 & 0.3703 & 0.8985 & 0.7514 \\
Mask-Diff (HINGE)   & \textbf{0.4871} & \textbf{0.3815} & \textbf{0.8956} & \textbf{0.7467}  \\ 
\midrule
\midrule
Hist-Affine-LN & 0.2285 & 0.1869 & 1.7018 & 1.0957     \\ 
SoftAdaLN (NoSoftNorm)  &  0.3962 & 0.2899 & 1.3680 & 0.9486 \\
SoftAdaLN (NoIdInit)   & 0.4160 & 0.3030 & 0.9644 & 0.8015  \\  
SoftAdaLN (Full)   & \textbf{0.4871} & \textbf{0.3815} & \textbf{0.8956} & \textbf{0.7467}  \\ 
\midrule
\midrule
ResNet-50    & 0.4455 & 0.3247 & 0.9811 & 0.8008 \\
UNI           & 0.4699 & 0.3642 & 0.9675 & 0.7884 \\
CONCH         & 0.4535 & 0.3557 & 1.0675 & 0.8151 \\
UNI + CONCH   & \textbf{0.4871} & \textbf{0.3815} & \textbf{0.8956} & \textbf{0.7467}  \\ 
\bottomrule
\end{tabular}
}
\caption{Ablations on Kidney. Component-wise analysis of HINGE variants on the Kidney (IU-F52) dataset.}
\label{tab:ablation-kidney}
\end{table}

\begin{table*}
\centering
\setlength{\tabcolsep}{4pt}
\resizebox{\textwidth}{!}
{
    \begin{tabular}{l|cccc|cccc|cccc}
    \toprule
    \multicolumn{1}{c|}{\multirow{2}{*}{Methods}} & \multicolumn{4}{c|}{cSCC} & \multicolumn{4}{c|}{Her2ST} & \multicolumn{4}{c}{Kidney}  \\ 
    \cmidrule(lr){2-5} \cmidrule(lr){6-9} \cmidrule(lr){10-13} 
    & PCC-50 $\uparrow$ & PCC-200 $\uparrow$ & MSE $\downarrow$ & MAE $\downarrow$  
    & PCC-50 $\uparrow$ & PCC-200 $\uparrow$ & MSE $\downarrow$ & MAE $\downarrow$  
    & PCC-50 $\uparrow$ & PCC-200 $\uparrow$ & MSE $\downarrow$ & MAE $\downarrow$ \\ 
    \midrule
    \midrule
    Linear  ($\zeta{=}1$) & 0.8728 & 0.7957 & 1.2067 & 0.8627    
            & 0.4922   & 0.3524   & 1.0802   & 0.8220   
            & {0.4817} & {0.3751} & {0.8986} & {0.7495}   \\
    Cosine  & 0.8641 & 0.7868 & 1.1760 & 0.8392    
            & 0.4652   & 0.3294   & 1.1293   & 0.8383   
            &  0.4579     &  0.3469      & 1.0026      &  0.7999      \\
    \(\zeta=0.5\)    & 0.8743 & 0.7964 & 1.2196 & 0.8576     
            & \textbf{0.5090}   & \textbf{0.3578}   & 1.1948   & 0.8619   
            &  0.4567     &   0.3485   &   0.9886  &  0.7948        \\
    \(\zeta=2\)  & 0.8657 & 0.7828 & 1.3044 & 0.8990    
            & 0.4913   & 0.3450   & 1.0377   & 0.7888   
            &  0.4773     &  0.3729    & 0.9504    &  0.7829       \\
    \(\zeta=\log_T G\)  &    \textbf{0.8755} & \textbf{0.8021} & \textbf{1.0096} & \textbf{0.7793}   
            & {0.4817} & {0.3751} & \textbf{0.8976} & \textbf{0.7445}     
            & \textbf{0.4871} & \textbf{0.3815} & \textbf{0.8956} & \textbf{0.7467}            \\
    \bottomrule
    \end{tabular}
}
\caption{Masking schedules. Results under different masking schedules on representative slices from the cSCC (P2\_ST\_rep3), Her2ST (A1), and Kidney (IU-F52) datasets.}
\label{tab:ablation-schedule}
\end{table*}

\subsection{Implementation Details}

Our approach is implemented using PyTorch (version 2.1.0) with Python 3.9, and models are trained on NVIDIA A800 GPUs with CUDA 12.1. We employ mixed precision training, utilizing PyTorch's native Automatic Mixed Precision (AMP) for computational efficiency. To ensure reproducibility, the random seed is consistently set at 42 across all experiments. The model is optimized using AdamW with a learning rate of $1 \times 10^{-4}$, weight decay of 0.0, and a global batch size of 32. We adopt a MultiStepLR scheduler with decay milestones at epochs [20, 30] and a decay factor of 0.2. The training process is capped at a maximum of 50 epochs, with an early stopping mechanism triggered if there is no improvement in validation MSE for 5 consecutive epochs after an initial validation warmup period of 15 epochs. To stabilize early-stage training, we implement a curriculum learning scheme where the first 5 epochs are restricted to mask ratios $\leq 20\%$ before exposing the model to the full diffusion schedule.

\section{Additional Visualization Results}
\begin{figure*}[t]
    \centering
    \includegraphics[width=\textwidth]{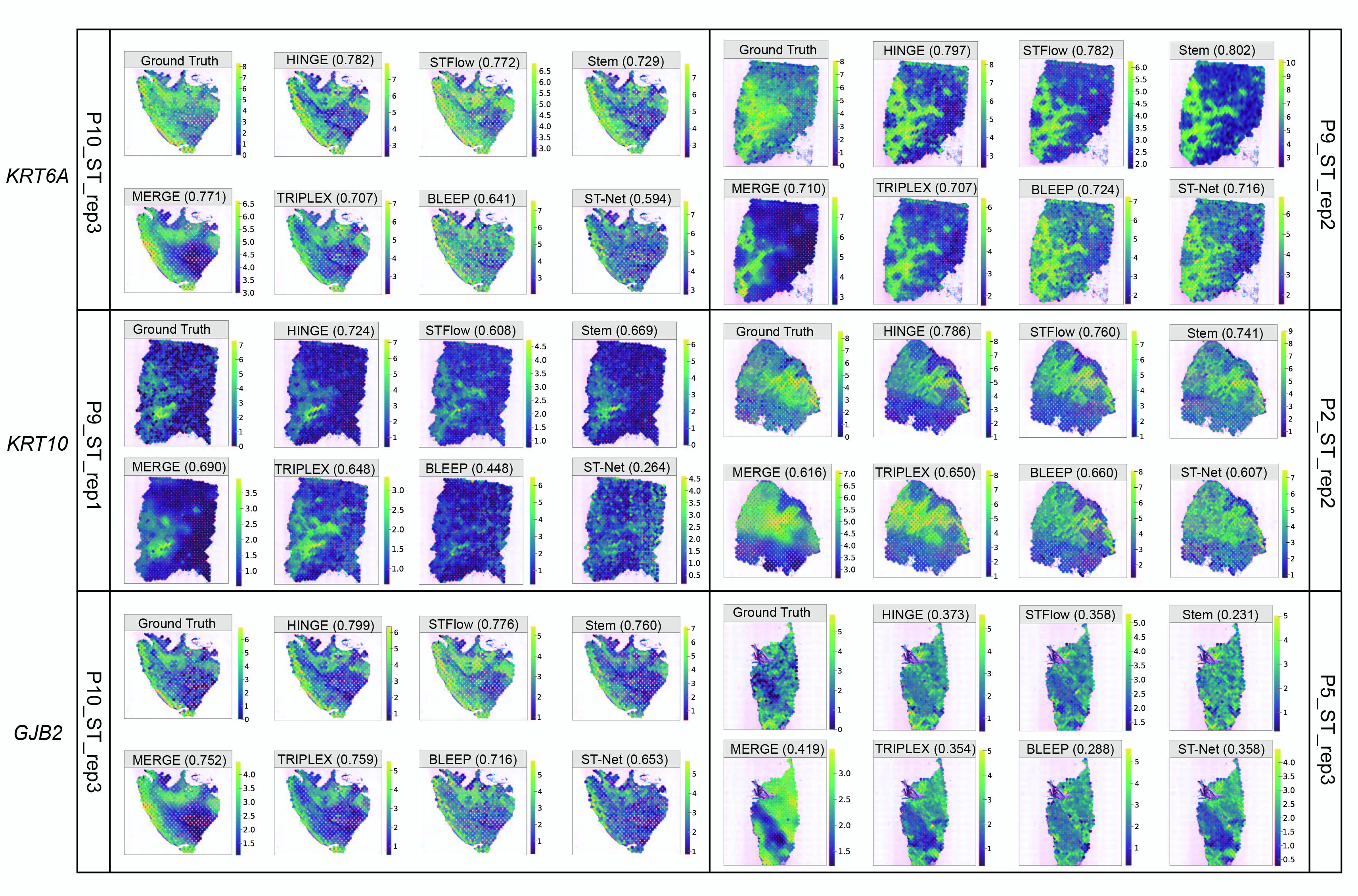}
    \caption{Spatial gene expression predictions for \textit{KRT6A}, \textit{KRT10}, and \textit{GJB2} on the \textbf{cSCC} dataset.}
    \label{fig:cscc}
\end{figure*}

\begin{figure*}[t]
    \centering
    \includegraphics[width=\textwidth]{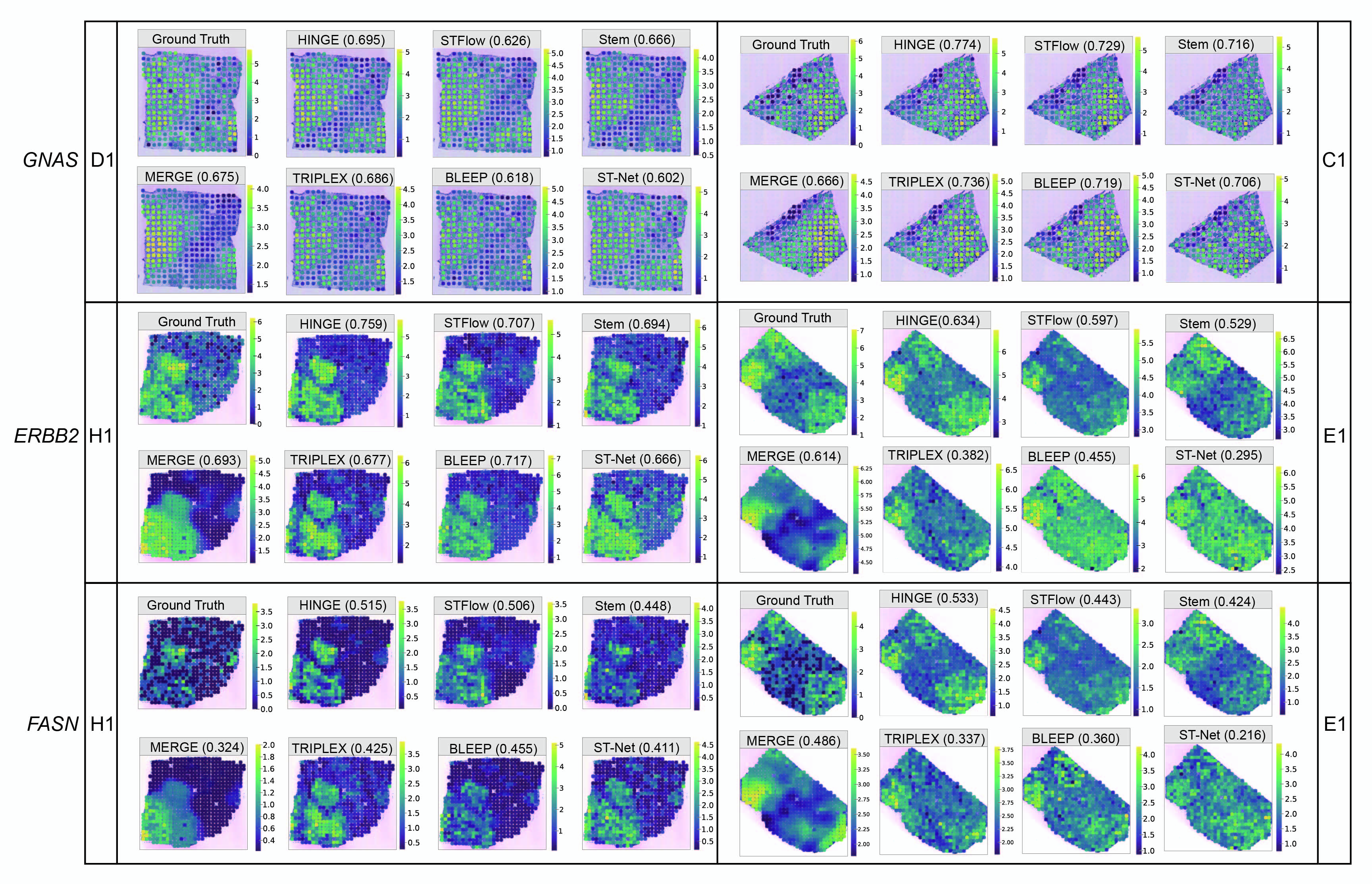}
    \caption{Spatial gene expression predictions for \textit{GNAS}, \textit{ERBB2}, and \textit{FASN} on the \textbf{Her2ST} dataset.}
    \label{fig:her2st}
\end{figure*}

\begin{figure*}[t]
    \centering
    \includegraphics[width=\textwidth]{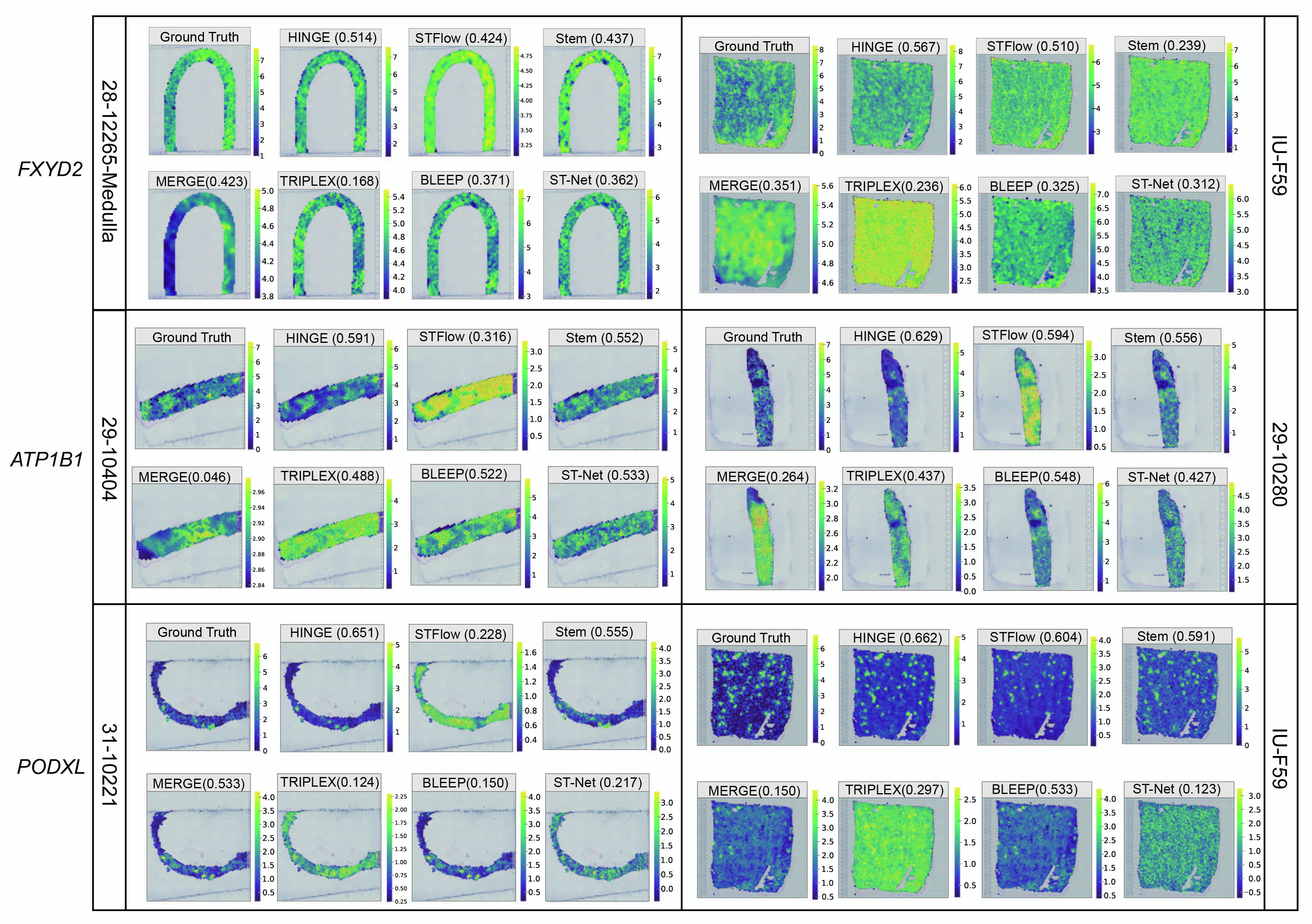}
    \caption{Spatial gene expression predictions for \textit{FXYD2}, \textit{ATP1B1}, and \textit{PODXL} on the \textbf{Kidney} dataset.}
    \label{fig:kidney}
\end{figure*}

In this section, we present additional visualizations of marker-gene spatial expression predictions across all three datasets used in our experiments (Figs.~\ref{fig:cscc}--\ref{fig:kidney}). These qualitative results complement the quantitative metrics in the main paper by providing a more fine-grained view of spatial localization patterns across datasets and tissue sections. 

For the cSCC dataset, Fig.~\ref{fig:cscc} shows spatial expression maps for \textit{KRT6A}, \textit{KRT10}, and \textit{GJB2} on multiple tissue sections. Beyond the P2\_ST\_rep3 slice shown in the main text, we include additional cSCC slices to illustrate how the predicted localization patterns of these markers behave across different sections rather than on a single example.

We follow the same protocol on the Her2ST and Kidney datasets. For Her2ST, Fig.~\ref{fig:her2st} visualizes predictions for \textit{GNAS}, \textit{ERBB2}, and \textit{FASN} on multiple tissue sections. For the Kidney dataset, Fig.~\ref{fig:kidney} shows \textit{FXYD2}, \textit{ATP1B1}, and \textit{PODXL} on representative slices. These examples are meant to complement the quantitative results in the main paper by visually inspecting whether the predicted marker-gene maps preserve the expected spatial structures and localization patterns across datasets and sections.

\begin{figure*}[t]
  \centering
  \includegraphics[width=0.9\linewidth]{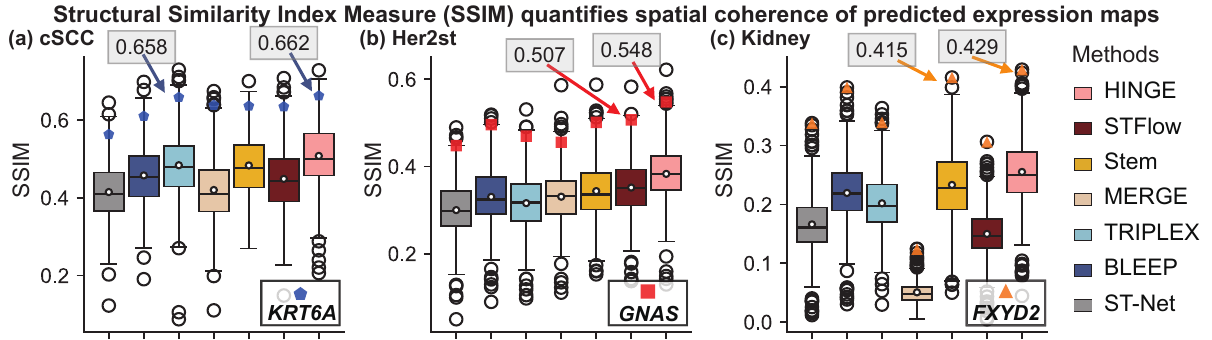}
  \caption{Gene-wise SSIM with marker genes highlighted.}
  \label{fig:ssim}
\end{figure*}

To further quantify spatial coherence beyond point-wise errors, we additionally report gene-wise structural similarity (SSIM) between the predicted and ground-truth 2D expression maps. Specifically, for each slice and each gene, we compute SSIM on the corresponding 2D spatial expression maps, and then aggregate the results per slice and finally average across slices. We highlight representative marker genes (\textit{KRT6A}, \textit{GNAS}, \textit{FXYD2}) in Fig.~\ref{fig:ssim} as references from cSCC, Her2ST, and Kidney, respectively. This SSIM-based evaluation provides a complementary perspective to MSE/MAE/PCC by emphasizing structural agreement of spatial patterns (e.g., contiguous regions and tissue-level organization) rather than purely per-spot deviations.

\begin{figure*}[t]
    \centering
    \includegraphics[width=\textwidth]{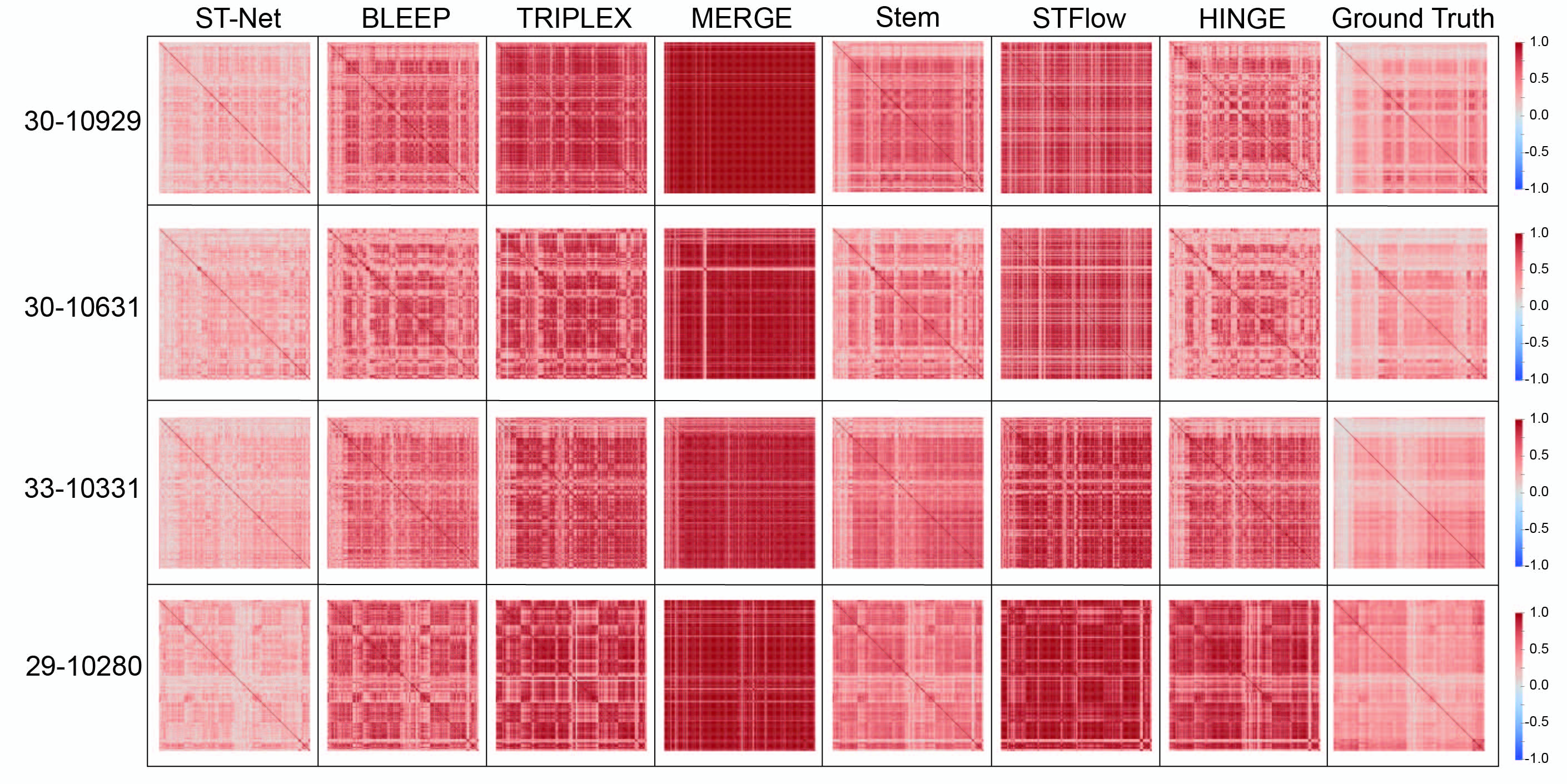}
    \caption{Gene--gene correlation heatmaps on the \textbf{Kidney} dataset.}
    \label{fig:gene-gene-cor}
\end{figure*}

In addition, for the Kidney dataset we visualize gene--gene correlation matrix heatmaps computed across multiple slices (Fig.~\ref{fig:gene-gene-cor}). These co-expression maps provide a complementary perspective to the marker-gene expression plots by highlighting gene--gene dependencies at the tissue level.

\section{Additional Ablation Studies}
\label{sec:supp:ablation}

We extend the ablation analysis from the main paper to the Her2ST (A1) and Kidney (IU-F52) datasets to verify that our design choices are not specific to cSCC. In all cases, we reuse the same protocol, metrics, and variant definitions as in the main-text ablations.

On the Her2ST (Table~\ref{tab:ablation-her2st}) and the Kidney dataset (Table~\ref{tab:ablation-kidney}), we report the same suite of ablations as in the main text. Each table includes the Scratch baseline and all variants that reuse the pre-trained CellFM backbone, compares Gaussian and masked diffusion objectives (including the HINGE schedule with curriculum), and lists conditioning designs such as Hist-Affine-LN and the full SoftAdaLN module, along with different histology encoders (UNI, CONCH, and UNI+CONCH).
The experiments follow the same protocol and metrics as the cSCC ablations in the main paper.

We further study different masking schedules on representative slices from the cSCC, Her2ST, and Kidney datasets (Table~\ref{tab:ablation-schedule}). In this experiment, we vary the forward masking schedule $\bar{\alpha}_t$ while keeping the rest of the setup fixed. We consider a linear schedule (\textbf{Linear}), a cosine schedule (\textbf{Cosine}), and two power-law schedules of the form $\bar{\alpha}_t = \bigl(1 - \tfrac{t}{T}\bigr)^{\zeta}$ with $\zeta \in \{0.5, 2\}$. In addition, we include a variant where the exponent is set to $\zeta = \log_T G$ for a prescribed global masking level $G$, which serves as the default schedule in our other experiments. Results are reported for all three datasets using the same evaluation metrics as in the main text. 

{With a full $T$-step run, we also evaluate the intermediate estimate $\hat{x}_0(t)$ at several $t$ values. MSE/MAE decrease and PCC increases until convergence, without late-step degradation (Fig.~\ref{fig:difftimesteps}(a)), suggesting progressive refinement rather than error accumulation under our progressive unmasking scheme.}
{In addition, fixing the trained model (with $T$), we vary the inference budget $K$ and report final metrics vs.\ $K$. A small $K$ already approaches the full-$T$ result, giving a clear quality--speed trade-off (Fig.~\ref{fig:difftimesteps}(b)).}

Finally, we examine the effect of the masking horizon $T$ on the same three datasets (Fig.~\ref{fig:ablation}). For each dataset, we fix the masking schedule and vary $T$ over several values, and then record the corresponding performance metrics. We visualize these results as curves of each metric versus $T$, providing a summary of how the choice of masking horizon interacts with our masked diffusion formulation on cSCC, Her2ST, and Kidney slices. Unless otherwise specified, we set $T = 50$ as the default masking horizon in our experiments.

\section{Inference efficiency} 
\begin{figure*}[t]
  \centering
  \includegraphics[width=0.8\linewidth]{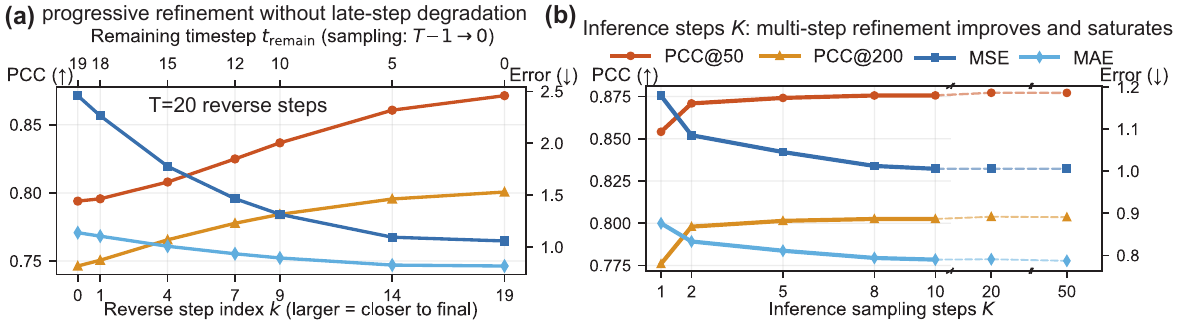}
   \caption{Step-wise analysis of masked diffusion sampling. }
   \label{fig:difftimesteps}
\end{figure*}

\begin{figure*}[t]
    \centering
    \includegraphics[width=0.9\textwidth]{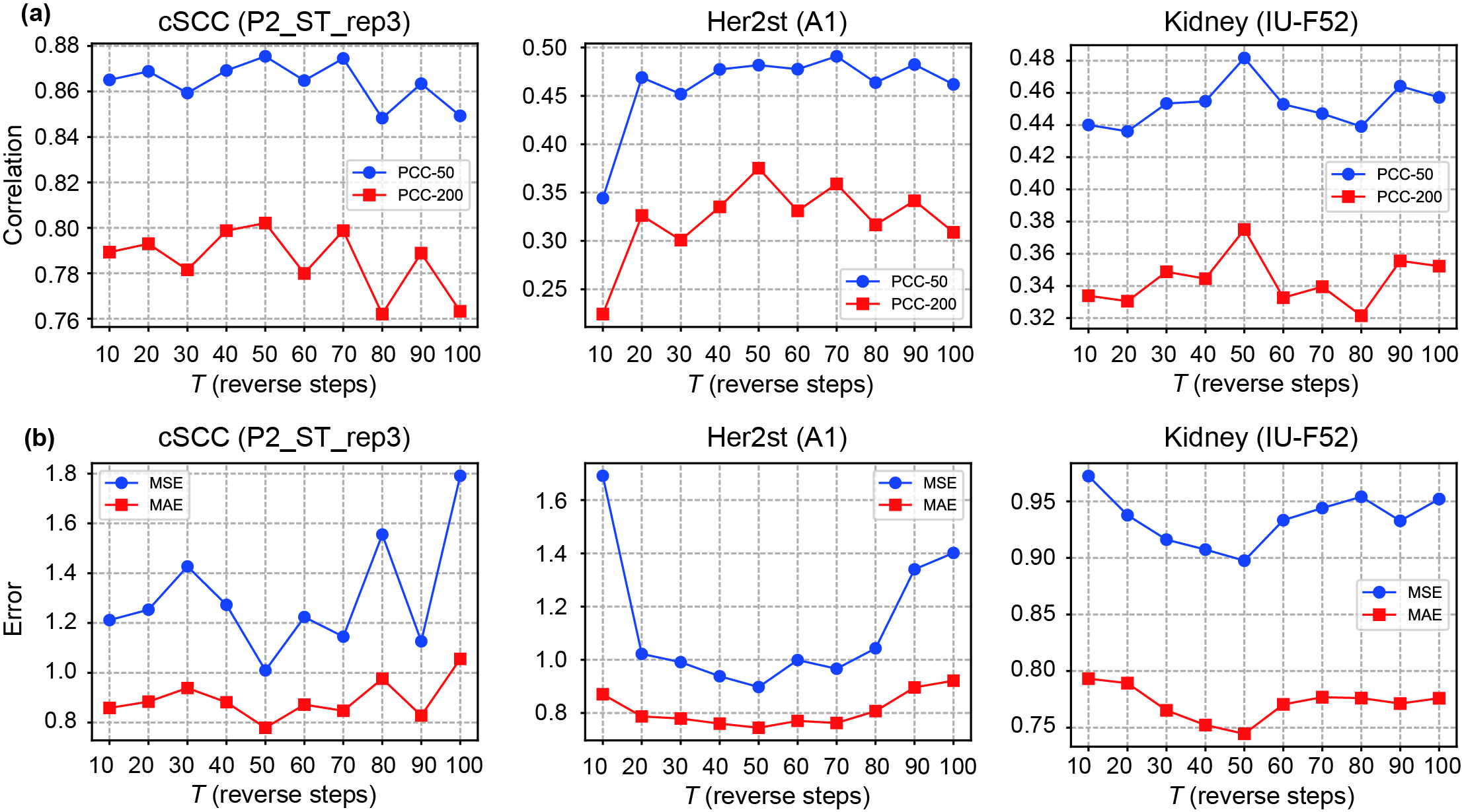}
    \caption{Masking horizon $T$. Evaluation metrics as functions of the masking horizon $T$ on representative slices from the cSCC, Her2ST, and Kidney datasets.}

    \label{fig:ablation}
\end{figure*}

\textbf{(i) Inference latency.} We add a runtime comparison of regression and generative baselines (\autoref{tab:comptcost}). 
At our default setting (HINGE (with-CellFM), \(T\)=50), throughput is 2.48 spots/s, comparable to Stem (2.77 spots/s) while achieving higher PCC. 
\textbf{(ii) Quality--speed trade-off (\(T\)).}
HINGE exposes the denoising steps as a practical test-time scaling: reducing \(T\) from 50 to 5 increases throughput by $\sim$11$\times$ (2.48$\rightarrow$26.46 spots/s, exceeding BLEEP in this setting) while keeping PCC nearly unchanged (0.877$\rightarrow$0.874; \autoref{tab:comptcost}, \autoref{fig:difftimesteps}). This shows HINGE reaches near-full accuracy with few steps, enabling practical inference throughput. 
\textbf{(iii) High-resolution case.} Inference is batched over spots and scales approximately linearly. Time@HR in \autoref{tab:comptcost} summarizes this regime, where some baselines are OOM while HINGE completes inference. 
\vspace{-2pt}
\begin{table}[t]
\centering
\label{tab:efficiency}
\setlength{\tabcolsep}{4pt}
\resizebox{\columnwidth}{!}{%
\begin{tabular}{l c c r r r| r}
\toprule
\textbf{Method} & \textbf{Type} & 
\textbf{Steps} &
\textbf{Time/slide (s)} $\downarrow$ &
\textbf{spots/s} $\uparrow$ &
\textbf{PCC@50} $\uparrow$ &
\textbf{Time@HR (s)} $\downarrow$ \\
\midrule
ST-Net & Reg. & 1 & 0.7162 & 890.82 & 0.739 & 13.97 \\
BLEEP & Reg. & 1 & 48.1776 & 13.24 & 0.785 & 991.29 \\
TRIPLEX & Reg. & 1 & 10.8633 & 58.45 & 0.805 & OOM \\
Stem & Gen. & 1000 & 230.1942 & 2.77 & 0.823 & 4138.51 \\
STFlow & Gen. & 10 & 0.2431 & 2624.83 & 0.692 & OOM \\
\midrule
HINGE (scGPT) & Gen. & 5 & 1.1643 & 547.97 & 0.806 & 37.3242 \\
HINGE (CellFM) & Gen. & 5 & 23.3428 & 26.46 & 0.874 & 453.06 \\
HINGE (CellFM) & Gen. & 50 & 257.7769 & 2.48 & 0.877 & 4411.02 \\
\bottomrule
\end{tabular}%
}
\caption{Inference efficiency (OOM: Out of Memory).
}
\label{tab:comptcost}
\end{table}

{
    \small
    \bibliographystyle{ieeenat_fullname}
    \bibliography{main}

@String(AAAI = {AAAI})

@article{lee2024pathomclip,
  title={{PathOmCLIP}: Connecting tumor histology with spatial gene expression via locally enhanced contrastive learning of Pathology and Single-cell foundation model},
  author={Lee, Yongju and Liu, Xinhao and Hao, Minsheng and Liu, Tianyu and Regev, Aviv},
  journal={bioRxiv},
  pages={2024--12},
  year={2024},
  publisher={Cold Spring Harbor Laboratory}
}

@article{yu2023stgcl,
  title={{stGCL}: A versatile cross-modality fusion method based on multi-modal graph contrastive learning for spatial transcriptomics},
  author={Yu, Na and Zhang, Daoliang and Zhang, Wei and Liu, Zhiping and Qiao, Xu and Wang, Chuanyuan and Zhao, Miaoqing and Chao, Baoting and Li, Wei and De Marinis, Yang and others},
  journal={bioRxiv},
  pages={2023--12},
  year={2023},
  publisher={Cold Spring Harbor Laboratory}
}

@article{fatemi2023feasibility,
  title={Feasibility of inferring spatial transcriptomics from single-cell histological patterns for studying colon cancer tumor heterogeneity},
  author={Fatemi, Michael Y and Lu, Yunrui and Sharma, Cyril and Feng, Eric and Azher, Zarif L and Diallo, Alos B and Srinivasan, Gokul and Rosner, Grace M and Pointer, Kelli B and Christensen, Brock C and others},
  journal={medRxiv},
  pages={2023--10},
  year={2023},
  publisher={Cold Spring Harbor Laboratory Press}
}

@article{luo2025deep,
  title={Deep learning in integrating spatial transcriptomics with other modalities},
  author={Luo, Jiajian and Fu, Jiye and Lu, Zuhong and Tu, Jing},
  journal={Briefings in Bioinformatics},
  volume={26},
  number={1},
  pages={bbae719},
  year={2025},
  publisher={Oxford University Press}
}

@article{yiu2025transformative,
  title={Transformative advances in single-cell omics: a comprehensive review of foundation models, multimodal integration and computational ecosystems},
  author={Yiu, Taylor and Chen, Bin and Wang, Haoyu and Feng, Genyi and Fu, Qiangqiang and Hu, Huijing},
  journal={Journal of Translational Medicine},
  volume={23},
  number={1},
  pages={1176},
  year={2025},
  publisher={Springer}
}

@article{neidlinger2025benchmarking,
  title={Benchmarking foundation models as feature extractors for weakly supervised computational pathology},
  author={Neidlinger, Peter and El Nahhas, Omar SM and Muti, Hannah Sophie and Lenz, Tim and Hoffmeister, Michael and Brenner, Hermann and van Treeck, Marko and Langer, Rupert and Dislich, Bastian and Behrens, Hans Michael and others},
  journal={Nature Biomedical Engineering},
  pages={1--11},
  year={2025},
  publisher={Nature Publishing Group UK London}
}

@inproceedings{zhengreparameterized,
  title={A Reparameterized Discrete Diffusion Model for Text Generation},
  author={Zheng, Lin and Yuan, Jianbo and Yu, Lei and Kong, Lingpeng},
  booktitle={First Conference on Language Modeling},
  year={2024}
}

@article{min2024multimodal,
  title={Multimodal contrastive learning for spatial gene expression prediction using histology images},
  author={Min, Wenwen and Shi, Zhiceng and Zhang, Jun and Wan, Jun and Wang, Changmiao},
  journal={Briefings in Bioinformatics},
  volume={25},
  number={6},
  pages={bbae551},
  year={2024},
  publisher={Oxford University Press}
}

@article{staahl2016visualization,
  title={Visualization and analysis of gene expression in tissue sections by spatial transcriptomics},
  author={St{\aa}hl, Patrik L and Salm{\'e}n, Fredrik and Vickovic, Sanja and Lundmark, Anna and Navarro, Jos{\'e} Fern{\'a}ndez and Magnusson, Jens and Giacomello, Stefania and Asp, Michaela and Westholm, Jakub O and Huss, Mikael and others},
  journal={Science},
  volume={353},
  number={6294},
  pages={78--82},
  year={2016},
  publisher={American Association for the Advancement of Science}
}

@article{jaume2024hest1k,
  title={Hest-1k: {A} dataset for spatial transcriptomics and histology image analysis},
  author={Jaume, Guillaume and Doucet, Paul and Song, Andrew and Lu, Ming Yang and Almagro P{\'e}rez, Cristina and Wagner, Sophia and Vaidya, Anurag and Chen, Richard and Williamson, Drew and Kim, Ahrong and others},
  journal={Advances in Neural Information Processing Systems},
  volume={37},
  pages={53798--53833},
  year={2024}
}

@article{lake2023kidney,
  title={An atlas of healthy and injured cell states and niches in the human kidney},
  author={Lake, Blue B and Menon, Rajasree and Winfree, Seth and Hu, Qiwen and Melo Ferreira, Ricardo and Kalhor, Kian and Barwinska, Daria and Otto, Edgar A and Ferkowicz, Michael and Diep, Dinh and others},
  journal={Nature},
  volume={619},
  number={7970},
  pages={585--594},
  year={2023},
  publisher={Nature Publishing Group UK London}
}

@article{schaar2024nicheformer,
  title={{Nicheformer}: a foundation model for single-cell and spatial omics},
  author={Schaar, Anna C and Tejada-Lapuerta, Alejandro and Palla, Giovanni and Gutgesell, Robert and Halle, Lennard and Minaeva, Mariia and Vornholz, Larsen and Dony, Leander and Drummer, Francesca and Bahrami, Mojtaba and others},
  journal={bioRxiv},
  pages={2024--04},
  year={2024},
  publisher={Cold Spring Harbor Laboratory}
}

@article{hao2024large,
  title={Large-scale foundation model on single-cell transcriptomics},
  author={Hao, Minsheng and Gong, Jing and Zeng, Xin and Liu, Chiming and Guo, Yucheng and Cheng, Xingyi and Wang, Taifeng and Ma, Jianzhu and Zhang, Xuegong and Song, Le},
  journal={Nature Methods},
  volume={21},
  number={8},
  pages={1481--1491},
  year={2024},
  publisher={Nature Publishing Group US New York}
}

@article{kleshchevnikov2022cell2location,
  title={Cell2location maps fine-grained cell types in spatial transcriptomics},
  author={Kleshchevnikov, Vitalii and Shmatko, Artem and Dann, Emma and Aivazidis, Alexander and King, Hamish W and Li, Tong and Elmentaite, Rasa and Lomakin, Artem and Kedlian, Veronika and Gayoso, Adam and others},
  journal={Nature Biotechnology},
  volume={40},
  number={5},
  pages={661--671},
  year={2022},
  publisher={Nature Publishing Group US New York}
}

@article{lu2024CONCH,
  title={A visual-language foundation model for computational pathology},
  author={Lu, Ming Y and Chen, Bowen and Williamson, Drew FK and Chen, Richard J and Liang, Ivy and Ding, Tong and Jaume, Guillaume and Odintsov, Igor and Le, Long Phi and Gerber, Georg and others},
  journal={Nature Medicine},
  volume={30},
  number={3},
  pages={863--874},
  year={2024},
  publisher={Nature Publishing Group US New York}
}

@article{chen2024UNI,
  title={Towards a general-purpose foundation model for computational pathology},
  author={Chen, Richard J and Ding, Tong and Lu, Ming Y and Williamson, Drew FK and Jaume, Guillaume and Song, Andrew H and Chen, Bowen and Zhang, Andrew and Shao, Daniel and Shaban, Muhammad and others},
  journal={Nature Medicine},
  volume={30},
  number={3},
  pages={850--862},
  year={2024},
  publisher={Nature Publishing Group US New York}
}

@article{ho2020DDPM,
  title={Denoising diffusion probabilistic models},
  author={Ho, Jonathan and Jain, Ajay and Abbeel, Pieter},
  journal={Advances in Neural Information Processing Systems},
  volume={33},
  pages={6840--6851},
  year={2020}
}

@article{lipman2022flow,
  title={Flow matching for generative modeling},
  author={Lipman, Yaron and Chen, Ricky TQ and Ben-Hamu, Heli and Nickel, Maximilian and Le, Matt},
  journal={arXiv preprint arXiv:2210.02747},
  year={2022}
}

@article{baek2025scfms,
  title={Single-cell foundation models: bringing artificial intelligence into cell biology},
  author={Baek, Seungbyn and Song, Kyungwoo and Lee, Insuk},
  journal={Experimental \& Molecular Medicine},
  pages={1--13},
  year={2025},
  publisher={Nature Publishing Group UK London}
}

@article{chen2025omic,
  title={A visual--omics foundation model to bridge histopathology with spatial transcriptomics},
  author={Chen, Weiqing and Zhang, Pengzhi and Tran, Tu N and Xiao, Yiwei and Li, Shengyu and Shah, Vrutant V and Cheng, Hao and Brannan, Kristopher W and Youker, Keith and Lai, Li and others},
  journal={Nature Methods},
  pages={1--15},
  year={2025},
  publisher={Nature Publishing Group US New York}
}

@article{theodoris2023transfer,
  title={Transfer learning enables predictions in network biology},
  author={Theodoris, Christina V and Xiao, Ling and Chopra, Anant and Chaffin, Mark D and Al Sayed, Zeina R and Hill, Matthew C and Mantineo, Helene and Brydon, Elizabeth M and Zeng, Zexian and Liu, X Shirley and others},
  journal={Nature},
  volume={618},
  number={7965},
  pages={616--624},
  year={2023},
  publisher={Nature Publishing Group UK London}
}

@article{cui2024scscgpt,
  title={{scGPT}: toward building a foundation model for single-cell multi-omics using generative AI},
  author={Cui, Haotian and Wang, Chloe and Maan, Hassaan and Pang, Kuan and Luo, Fengning and Duan, Nan and Wang, Bo},
  journal={Nature Methods},
  volume={21},
  number={8},
  pages={1470--1480},
  year={2024},
  publisher={Nature Publishing Group US New York}
}

@article{wang2025scgpt,
  title={{scGPT-spatial}: Continual pretraining of single-cell foundation model for spatial transcriptomics},
  author={Wang, Chloe and Cui, Haotian and Zhang, Andrew and Xie, Ronald and Goodarzi, Hani and Wang, Bo},
  journal={bioRxiv},
  pages={2025--02},
  year={2025},
  publisher={Cold Spring Harbor Laboratory}
}

@article{zeng2025cellfm,
  title={{CellFM}: a large-scale foundation model pre-trained on transcriptomics of 100 million human cells},
  author={Zeng, Yuansong and Xie, Jiancong and Shangguan, Ningyuan and Wei, Zhuoyi and Li, Wenbing and Su, Yun and Yang, Shuangyu and Zhang, Chengyang and Zhang, Jinbo and Fang, Nan and others},
  journal={Nature Communications},
  volume={16},
  number={1},
  pages={4679},
  year={2025},
  publisher={Nature Publishing Group UK London}
}

@inproceedings{shi2025multigo,
  title={Multi-modal Topology-embedded Graph Learning for Spatially Resolved Genes Prediction from Pathology Images with Prior Gene Similarity Information},
  author={Shi, Hang and Chi, Changxi and Wan, Peng and Zhang, Daoqiang and Shao, Wei},
  booktitle={Proceedings of the Computer Vision and Pattern Recognition Conference},
  pages={20810--20819},
  year={2025}
}

@inproceedings{
zhao2025stofm,
title={{ST}o{FM}: a Multi-scale Foundation Model for Spatial Transcriptomics},
author={Suyuan Zhao and YIZHEN LUO and Ganbo Yang and Yan Zhong and Hao Zhou and Zaiqing Nie},
booktitle={Forty-second International Conference on Machine Learning},
year={2025},
url={https://openreview.net/forum?id=PQx66EJUu0}
}

@inproceedings{wang2025m2ost,
  title={{M2OST}: Many-to-one Regression for Predicting Spatial Transcriptomics from Digital Pathology Images},
  author={Wang, Hongyi and Du, Xiuju and Liu, Jing and Ouyang, Shuyi and Chen, Yen-Wei and Lin, Lanfen},
  booktitle={Proceedings of the AAAI Conference on Artificial Intelligence},
  pages={7709--7717},
  year={2025}
}

@article{li2024gene,
  title={Gene expression prediction from histology images via hypergraph neural networks},
  author={Li, Bo and Zhang, Yong and Wang, Qing and Zhang, Chengyang and Li, Mengran and Wang, Guangyu and Song, Qianqian},
  journal={Briefings in Bioinformatics},
  volume={25},
  number={6},
  pages={bbae500},
  year={2024},
  publisher={Oxford University Press}
}

@article{zeng202hist2st,
  title={Spatial transcriptomics prediction from histology jointly through transformer and graph neural networks},
  author={Zeng, Yuansong and Wei, Zhuoyi and Yu, Weijiang and Yin, Rui and Yuan, Yuchen and Li, Bingling and Tang, Zhonghui and Lu, Yutong and Yang, Yuedong},
  journal={Briefings in Bioinformatics},
  volume={23},
  number={5},
  year={2022},
  publisher={Oxford Academic}
}

@article{pang2021hisTogene,
  title={Leveraging information in spatial transcriptomics to predict super-resolution gene expression from histology images in tumors},
  author={Pang, Minxing and Su, Kenong and Li, Mingyao},
  journal={bioRxiv},
  pages={2021--11},
  year={2021},
  publisher={Cold Spring Harbor Laboratory}
}

@inproceedings{chung2024TRIPLEX,
  title={Accurate spatial gene expression prediction by integrating multi-resolution features},
  author={Chung, Youngmin and Ha, Ji Hun and Im, Kyeong Chan and Lee, Joo Sang},
  booktitle={Proceedings of the IEEE/CVF Conference on Computer Vision and Pattern Recognition},
  pages={11591--11600},
  year={2024}
}

@article{he2020stnet,
  title={Integrating spatial gene expression and breast tumour morphology via deep learning},
  author={He, Bryan and Bergenstr{\aa}hle, Ludvig and Stenbeck, Linnea and Abid, Abubakar and Andersson, Alma and Borg, {\AA}ke and Maaskola, Jonas and Lundeberg, Joakim and Zou, James},
  journal={Nature Biomedical Engineering},
  volume={4},
  number={8},
  pages={827--834},
  year={2020},
  publisher={Nature Publishing Group UK London}
}

@article{andersson2020her2st,
  title={Spatial deconvolution of {HER2}-positive breast tumors reveals novel intercellular relationships},
  author={Andersson, Alma and Larsson, Ludvig and Stenbeck, Linnea and Salm{\'e}n, Fredrik and Ehinger, Anna and Wu, Sunny and Al-Eryani, Ghamdan and Roden, Daniel and Swarbrick, Alex and Borg, {\AA}ke and others},
  journal={bioRxiv},
  pages={2020--07},
  year={2020},
  publisher={Cold Spring Harbor Laboratory}
}

@article{ji2020SCC,
  title={Multimodal analysis of composition and spatial architecture in human squamous cell carcinoma},
  author={Ji, Andrew L and Rubin, Adam J and Thrane, Kim and Jiang, Sizun and Reynolds, David L and Meyers, Robin M and Guo, Margaret G and George, Benson M and Mollbrink, Annelie and Bergenstr{\aa}hle, Joseph and others},
  journal={Cell},
  volume={182},
  number={2},
  pages={497--514},
  year={2020},
  publisher={Elsevier}
}

@article{han2025reusability,
  title={Reusability report: Exploring the transferability of self-supervised learning models from single-cell to spatial transcriptomics},
  author={Han, Chuangyi and Lin, Senlin and Wang, Zhikang and Cui, Yan and Zou, Qi and Yuan, Zhiyuan},
  journal={Nature Machine Intelligence},
  pages={1--15},
  year={2025},
  publisher={Nature Publishing Group UK London}
}

@article{xie2023BLEEP,
  title={Spatially resolved gene expression prediction from histology images via bi-modal contrastive learning},
  author={Xie, Ronald and Pang, Kuan and Chung, Sai and Perciani, Catia and MacParland, Sonya and Wang, Bo and Bader, Gary},
  journal={Advances in Neural Information Processing Systems},
  volume={36},
  pages={70626--70637},
  year={2023}
}

@inproceedings{ganguly2025merge,
  title={{MERGE}: Multi-faceted Hierarchical Graph-based GNN for Gene Expression Prediction from Whole Slide Histopathology Images},
  author={Ganguly, Aniruddha and Chatterjee, Debolina and Huang, Wentao and Zhang, Jie and Yurovsky, Alisa and Johnson, Travis Steele and Chen, Chao},
  booktitle={Proceedings of the Computer Vision and Pattern Recognition Conference},
  pages={15611--15620},
  year={2025}
}

@inproceedings{zhu2025Stem,
    title={Diffusion Generative Modeling for Spatially Resolved Gene Expression Inference from Histology Images},
    author={Sichen Zhu and Yuchen Zhu and Molei Tao and Peng Qiu},
    booktitle={The Thirteenth International Conference on Learning Representations},
    year={2025}
}

@inproceedings{huang2025STFlow,
    title={Scalable Generation of Spatial Transcriptomics from Histology Images via Whole-Slide Flow Matching},
    author={Tinglin Huang and Tianyu Liu and Mehrtash Babadi and Wengong Jin and Rex Ying},
    booktitle={Forty-second International Conference on Machine Learning},
    year={2025}
}

@inproceedings{huang2017densely,
  title={Densely connected convolutional networks},
  author={Huang, Gao and Liu, Zhuang and Van Der Maaten, Laurens and Weinberger, Kilian Q},
  booktitle={Proceedings of the IEEE conference on computer vision and pattern recognition},
  pages={4700--4708},
  year={2017}
}

@article{min2025spamask,
  title={{SpaMask}: Dual masking graph autoencoder with contrastive learning for spatial transcriptomics},
  author={Min, Wenwen and Fang, Donghai and Chen, Jinyu and Zhang, Shihua},
  journal={PLOS Computational Biology},
  volume={21},
  number={4},
  pages={e1012881},
  year={2025},
  publisher={Public Library of Science San Francisco, CA USA}
}

@article{fang2025spacross,
  title={{SpaCross} deciphers spatial structures and corrects batch effects in multi-slice spatially resolved transcriptomics},
  author={Fang, Donghai and Min, Wenwen},
  journal={Communications Biology},
  volume={8},
  number={1},
  pages={1393},
  year={2025},
  publisher={Nature Publishing Group UK London}
}

@article{niu2025spabatch,
  title={{SpaBatch}: Deep Learning-Based Cross-Slice Integration and 3D Spatial Domain Identification in Spatial Transcriptomics},
  author={Niu, Jinyun and Fang, Donghai and Chen, Jinyu and Xiong, Yi and Liu, Juan and Min, Wenwen},
  journal={Advanced Science},
  volume={12},
  number={44},
  pages={e09090},
  year={2025},
  publisher={Wiley Online Library}
}

@inproceedings{xue2025inferring,
  title={Inferring Super-Resolved Gene Expression by Integrating Histology Images and Spatial Transcriptomics with {HISTEX}},
  author={Xue, Shuailin and Wang, Changmiao and Fan, Xiaomao and Min, Wenwen},
  booktitle={International Conference on Medical Image Computing and Computer-Assisted Intervention},
  pages={296--306},
  year={2025},
  organization={Springer}
}
}

\newpage

\clearpage
\newpage

\end{document}